\documentclass[11pt]{article}
\usepackage[final]{rbs}

\usepackage{amsmath,amsfonts,bm}
\usepackage{xspace}
\usepackage{xcolor}

\def\eqref#1{equation~\ref{#1}}

\def\1{\bm{1}}

\DeclareMathAlphabet{\mathsfit}{\encodingdefault}{\sfdefault}{m}{sl}
\SetMathAlphabet{\mathsfit}{bold}{\encodingdefault}{\sfdefault}{bx}{n}

\usepackage{hyperref}
\usepackage{url}
\usepackage{graphicx}
\usepackage{float}
\usepackage{booktabs}
\usepackage{multirow}
\usepackage{subcaption}
\usepackage{wrapfig}
\usepackage{xcolor}
\usepackage{xspace}
\usepackage{amsmath}
\usepackage{enumitem}
\usepackage{booktabs}
\usepackage{multirow}
\usepackage{array}

\title{Motion Beyond Pixels: Learning Future Visual Representations for Object-Centric Dynamics}

\author{\textbf{Chengzhi Zhao}$^{1,2,*}$,
\textbf{Juelin Li}$^{1,3,*}$,
\textbf{Zhennan Wang}$^{1,\dagger}$,
\textbf{Sheng Wang}$^{1,\dagger}$,
\textbf{Zhuoman Liu}$^{1}$,\\
\textbf{Yuhao Lu}$^{1}$,
\textbf{Jiongxu Chen}$^{1}$,
\textbf{Haitang Feng}$^{1}$,
\textbf{Yanjie Li}$^{2}$,
\textbf{Lin Shao}$^{1,4,\ddagger}$\\[0.35em]
$^{1}$RoboScience \quad
$^{2}$Harbin Institute of Technology, Shenzhen\\
$^{3}$Sun Yat-sen University \quad
$^{4}$National University of Singapore
}

\begin{document}

\maketitle
\begingroup
\makeatletter
\renewcommand{\@makefntext}[1]{\noindent#1}
\makeatother
\renewcommand{\thefootnote}{}
\renewcommand{\theHfootnote}{rbscredits}
\begin{NoHyper}
\footnotetext{\textsuperscript{*} Equal contribution.\quad \textsuperscript{\ensuremath{\dagger}} Project lead.\quad \textsuperscript{\ensuremath{\ddagger}} Corresponding author.}
\end{NoHyper}
\endgroup

\begin{abstract}
Many embodied world models predict future states in latent spaces optimized for RGB reconstruction, which do not explicitly prioritize object-level semantics and geometry relevant to motion generation. 
We present MotionRep, a visual representation world model that jointly generates future DINO representations and 3D object flow, coupling semantic and geometric information with object-centric dynamics.
To make high-dimensional DINO features amenable to generative modeling, we introduce DINO-VAE, which combines temporal and channel compression with latent regularization to produce compact latents while preserving semantic and geometric information.
Conditioned on an initial observation, object query points, and a language instruction, a unified diffusion transformer predicts future visual representations and 3D object flow through joint attention.
MotionRep employs asymmetric modality-specific noise schedules to keep the visual stream ahead of 3D object flow during joint denoising, allowing progressively refined future visual representations to guide flow generation.
Experiments on simulated and real-world manipulation benchmarks demonstrate that MotionRep outperforms pixel-based baselines in 3D object flow accuracy and generalization to unseen scenes, while requiring only three denoising steps for inference.
These results support predictive visual representations as an effective future-state space for object-centric dynamics modeling.

{\sffamily\hypersetup{hidelinks}
\textbf{Project page:}\enspace
\href{https://motionrep.github.io/}
{\textcolor{rbsGray}{https://motionrep.github.io/}}}
\end{abstract}
\section{Introduction}

Effective robotic manipulation requires reasoning beyond the current observation to anticipate future object dynamics.
3D object flow explicitly represents the trajectories of object points in 3D space, providing a natural description of future physical dynamics~\citep{yuan2024general}. 
Recent 3D object flow world models incorporate future visual state modeling to support motion generation, making the choice of visual representation a critical design decision~\citep{li2026novaflow,lu2026worldflow}.

Existing 3D object flow world models commonly predict future states in pixel-reconstruction-oriented latent spaces~\citep{li2026novaflow,lu2026worldflow}, which preserve substantial low-level texture and appearance details but do not explicitly prioritize object semantics and spatial geometry relevant to motion prediction~\citep{yu2025repa,yao2025vavae}. Modeling such complex appearance details can increase inference cost and limit generalization to unseen scenes.
In contrast, visual representation world models predict future states rich in semantic and geometric information~\citep{zhou2025dinowm,assran2025vjepa2,guo2026v}.
However, these approaches primarily focus on visual representation prediction, while jointly modeling visual representations and 3D object flow remains underexplored.

\begin{figure*}[t]
    \centering
    \includegraphics[width=0.95\linewidth]{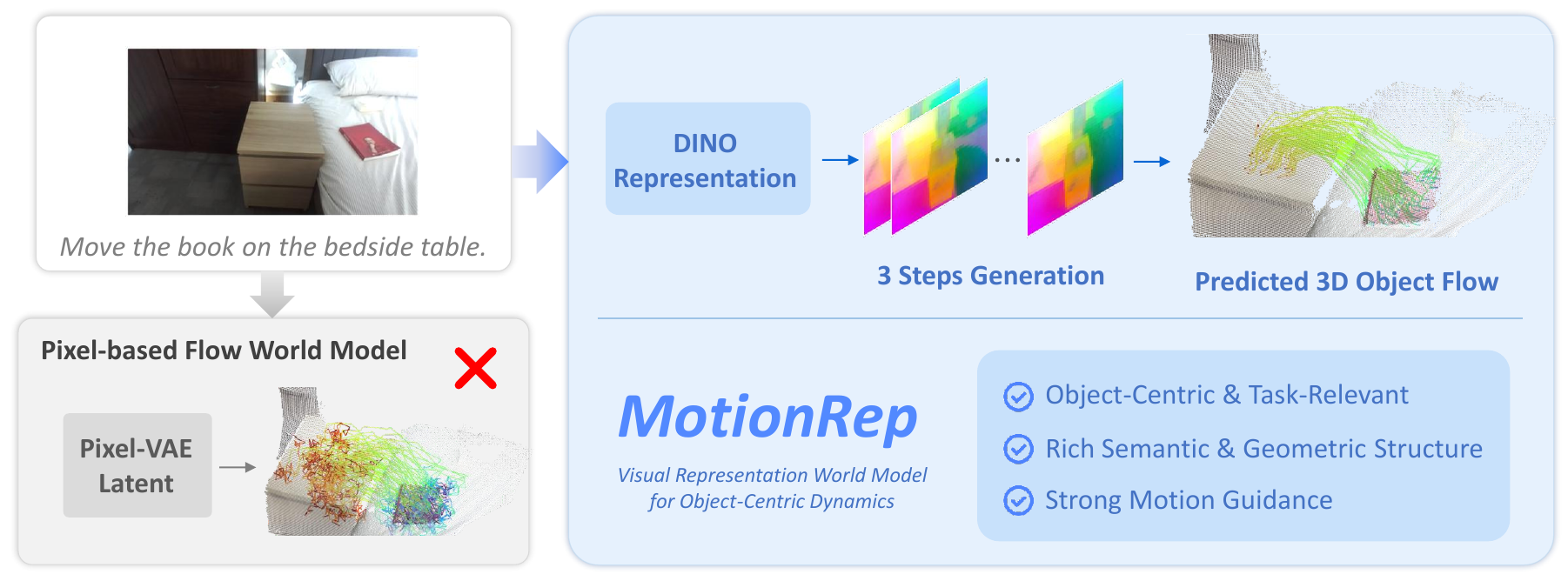}
    \caption{\textbf{Future-state representation matters for 3D object flow generation.} 
    Pixel-based flow world models generate future RGB frames in appearance-oriented VAE latent spaces, emphasizing visual details over motion-relevant semantics and geometry.
    In contrast, MotionRep models future states in the DINO representation space, leveraging rich semantic and geometric information for more accurate 3D object flow generation with fewer denoising steps.}
    \label{fig:overview}
\end{figure*}

To address this challenge, we introduce MotionRep, a visual representation world model for object-centric dynamics that jointly generates future DINO representations and 3D object flow, modeling future visual states in the DINO representation space rather than pixel-reconstruction-oriented latent spaces (Fig.~\ref{fig:overview}).
Although DINO representations capture rich semantic and geometric information~\citep{simeoni2025dinov3}, their high dimensionality and unregularized feature distributions limit direct diffusion modeling~\mbox{\citep{zheng2026rae,vfmvae2026}}.
We therefore introduce DINO-VAE, which combines temporal and channel compression with latent regularization to produce compact generative latents while preserving informative feature structure.
Building on this representation, MotionRep generates future visual representations and 3D object flow using a diffusion transformer, enabling direct interaction between visual states and object dynamics through joint attention.
Furthermore, we introduce decoupled timestep sampling and modality-specific noise schedules, allowing future visual representations to be progressively denoised ahead of 3D object flow and continuously guide motion generation in a latent-forcing manner.

Experiments on simulated and real-world benchmarks demonstrate that MotionRep achieves state-of-the-art 3D object flow prediction performance, generalizes to unseen scenes, and enables efficient generation with only three sampling steps.
Representation analyses show that DINO-VAE preserves semantic and geometric information in compressed latents. 
Ablation studies further validate the contributions of DINO-VAE, future DINO generation, modality-specific noise scheduling, and representation world model pretraining to 3D object flow generation. 
Conditional inference experiments demonstrate that MotionRep supports both visual-conditioned inverse dynamics and flow-conditioned forward dynamics within the same model, revealing bidirectional dependencies between future visual representations and 3D object motion.

Our main contributions are summarized as follows:
\begin{itemize}
    \item We demonstrate the value of explicitly predicting future visual representations for 3D object flow generation, and introduce MotionRep to jointly model future DINO representations and object motion.

    \item We introduce DINO-VAE, which makes dense DINO features amenable to generative modeling through temporal and channel compression and latent regularization, while preserving semantic and geometric information.

    \item We develop an asymmetric denoising strategy with decoupled timesteps and modality-specific noise schedules,  enabling representation-guided flow generation and bidirectional conditional dynamics modeling.
\end{itemize}

\section{Related Work}

\begin{figure*}[t]
    \centering
    \includegraphics[width=0.95\linewidth]{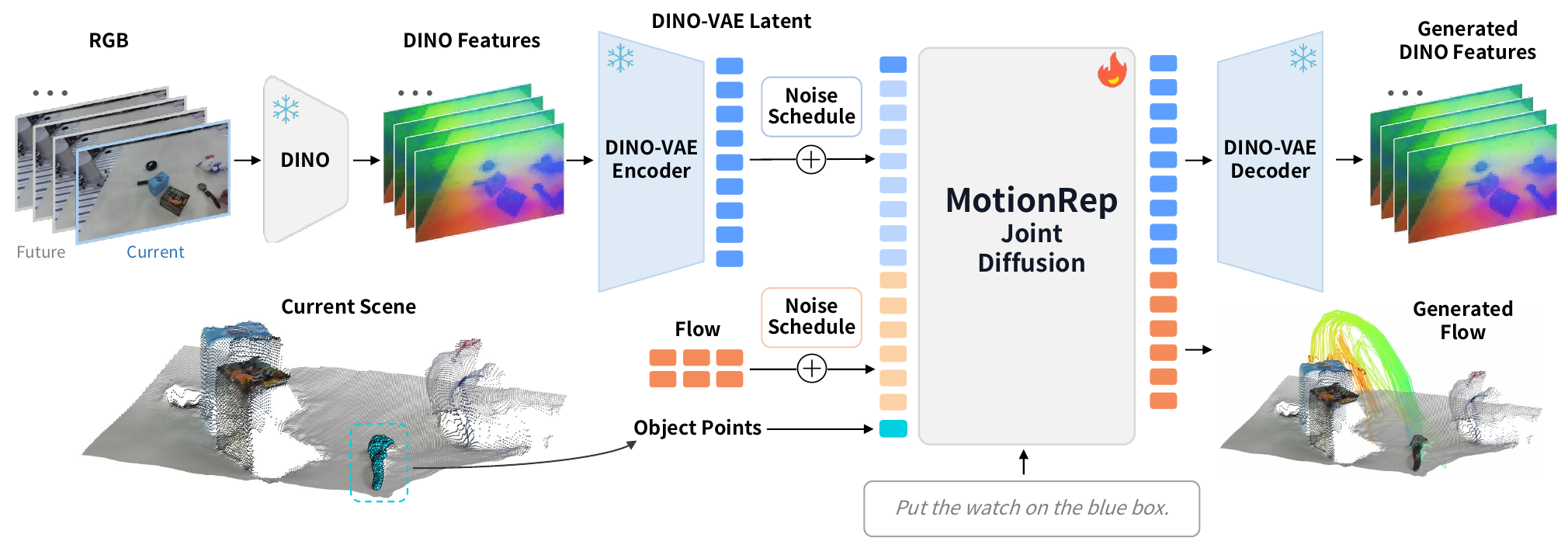}
    \caption{\textbf{Overview of MotionRep.}
    Given an initial RGB observation, 3D object query points, and a language instruction, MotionRep jointly generates future DINO representations and 3D object flow within a unified diffusion transformer. DINO-VAE compresses high-dimensional DINO features into compact visual latents, while modality-specific noise schedules enable asymmetric denoising, allowing progressively refined visual representations to guide 3D object flow generation.}
    \label{fig:architecture}
\end{figure*}

\paragraph{3D Object Flow World Models.}

Vision-language-action (VLA) models directly predict robot actions from multimodal observations~\citep{kim2024openvla,black2024pi0,brohan2023rt2}, while world action models (WAMs) incorporate future state prediction to explicitly model environment dynamics~\citep{du2023unipi,li2026lingbotva,tian2025seer,ye2026world,hu2024video}. However, video generation world models primarily focus on visual scene evolution and often lack explicit representations of 3D object motion. 
In contrast, 3D object flow captures object dynamics through 3D point trajectories, providing explicit geometric and motion information for robotic manipulation~\citep{yuan2024general,zhi20253dflowaction,lin2026roboflow4d}.
Existing 3D object flow generation methods follow two paradigms. Two-stage approaches generate future videos and recover 3D trajectories using point trackers~\citep{li2026novaflow,karaev2024cotracker,lu2026track4world,karhade2026any4d}, suffering from accumulated generation and tracking errors. Single-stage approaches directly generate 3D object flow~\citep{lee2026mu0,lee2026tracegen,zhang2026molmomotion,li2026egowam}, with some methods further jointly generating future videos~\citep{lu2026worldflow}. However, direct flow generation often lacks explicit future state guidance, while joint video-flow models typically rely on pixel-oriented VAE latents, which may not fully capture semantic and geometric information. In contrast, MotionRep jointly generates future visual representations and 3D object flow in the semantically and geometrically rich DINO representation space, providing more informative future state guidance for 3D motion generation.

\paragraph{Visual Representation World Models.}

Most video world models operate in pixel VAE latent spaces optimized for visual reconstruction~\citep{rombach2022ldm,yang2025cogvideox,wan2025,blattmann2023svd,blattmann2023videoldm}, devoting substantial capacity to low-level appearance details that may be less relevant to dynamics modeling. In contrast, self-supervised visual models such as DINO and V-JEPA learn representations rich in semantic and spatial information~\citep{caron2021dino,oquab2023dinov2,amir2021deepvit,bardes2024vjepa,simeoni2025dinov3,assran2025vjepa2}. Recent studies have explored these representations for generative modeling, either to improve optimization~\citep{yu2025repa,yao2025vavae} or as direct generation spaces~\citep{zheng2026rae,singh2026improved,guo2026v,xie2026videorae}, demonstrating faster convergence and improved generation quality.
However, much of the progress in representation-based generation has been demonstrated on images~\citep{yu2025repa,yao2025vavae,zheng2026rae}, while high-dimensional features~\citep{zheng2026rae} and temporal redundancy pose additional challenges for video generation. Meanwhile, world models operating in visual representation spaces primarily predict future visual states~\citep{zhou2025dinowm,assran2025vjepa2,boduljak2025vfmf}, leaving their potential for guiding 3D object motion generation underexplored. Unlike approaches that generate directly in high-dimensional representation spaces, MotionRep introduces DINO-VAE to construct a compact generative latent space and jointly models future visual representations and 3D object flow, leveraging rich semantic and geometric information to guide 3D motion generation.

\section{Method}

\subsection{Problem Formulation and Overview}
\label{sec:Overview}

\noindent\textbf{Problem Formulation.}
Given the initial DINO representation $X_0$, the 3D point set $P_0$ of the target object, and a task description $l$, MotionRep jointly models future DINO representations and 3D object flow within a unified framework. We treat 3D object flow as a high-level action representation. For the $p$-th point, its object flow at time $t$ is defined as the 3D displacement relative to its initial position $F_{t,p} = P_{t,p} - P_{0,p}$.
Let $X_{1:T}$ denote the future DINO feature sequence. MotionRep jointly generates future visual representations and 3D object flow as
\begin{equation}
(\hat{X}_{1:T}, \hat{F}_{1:T})
\sim
\pi_{\theta}^{\mathrm{}}
\left(
X_{1:T}, F_{1:T}
\mid
X_0, P_0, l
\right).
\end{equation}

With the same bidirectional conditional modeling framework, MotionRep also supports visual-conditioned inverse dynamics and flow-conditioned forward dynamics:
\begin{equation}
\hat{F}_{1:T}
\sim
\pi_{\theta}^{\mathrm{IDM}}
\left(
F_{1:T}
\mid
X_{0:T}, P_0, l
\right),
\end{equation}
\begin{equation}
\hat{X}_{1:T}
\sim
\pi_{\theta}^{\mathrm{FDM}}
\left(
X_{1:T}
\mid
X_0, F_{1:T}, P_0, l
\right).
\end{equation}

\noindent\textbf{Overview.}
The framework of MotionRep is illustrated in Fig.~\ref{fig:architecture}. First, we use a frozen DINO encoder to extract visual features with rich semantic and spatial structure, and employ DINO-VAE to compress the high-dimensional DINO features into compact latent representations suitable for generative modeling (Sec.~\ref{sec:dino_vae}). MotionRep then jointly models future visual representations and 3D object flow within a unified generative framework (Sec.~\ref{sec:joint_modeling}). Finally, we introduce modality-specific timestep sampling and noise schedules to model cross-modal conditional dependencies under different noise levels (Sec.~\ref{sec:joint_modeling}).

\subsection{Visual Representation Compression}
\label{sec:dino_vae}

\begin{figure}[t]
    \centering
    \includegraphics[width=0.8\linewidth]{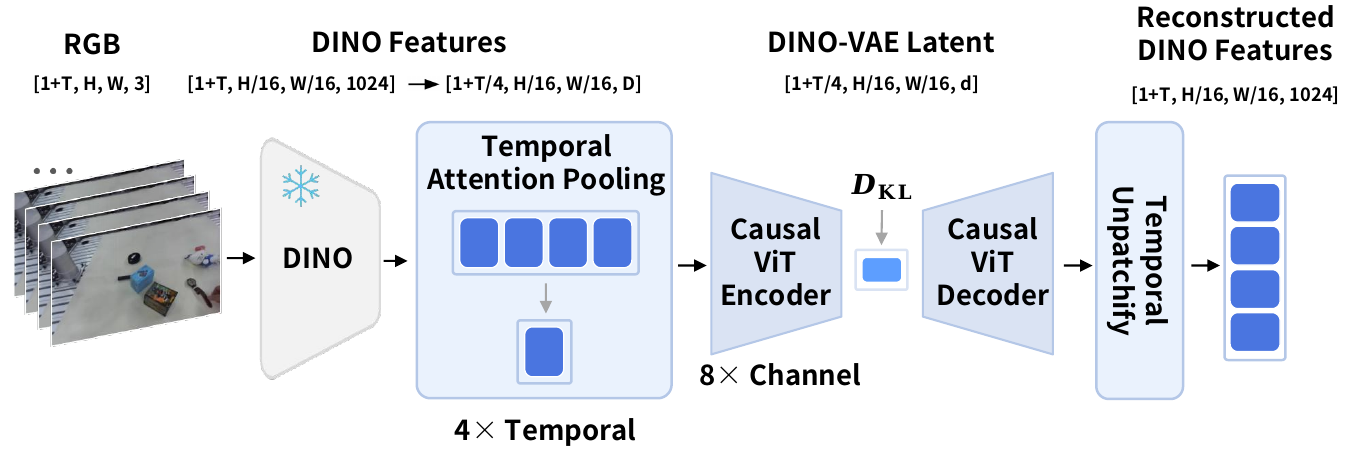}
    \caption{\textbf{Overview of DINO-VAE.}
    DINO-VAE compresses high-dimensional DINO features through temporal attention pooling and a causal ViT encoder, achieving $4\times$ temporal and $8\times$ channel compression. KL regularization facilitates generative modeling in the latent space, while a causal ViT decoder reconstructs the DINO features.}
    \label{fig:dino-vae}
    \vspace{-8pt}
\end{figure}

Visual representation models such as DINO~\citep{simeoni2025dinov3} provide rich object-level semantics and spatial-geometric structure, yet their raw features are not tailored to generative modeling. 
As shown in Fig.~\ref{fig:dino-vae}, we therefore introduce DINO-VAE to compress DINO features into a compact latent space while preserving their semantic and geometric structure.

\noindent\textbf{Temporal and Channel Compression.}
Given an input video $V \in \mathbb{R}^{(1+T) \times H \times W \times 3}$, we employ DINOv3-L to extract patch-level features $X \in \mathbb{R}^{(1+T) \times \frac{H}{16} \times \frac{W}{16} \times 1024}$.
Since frame-wise DINO features exhibit substantial temporal redundancy, modeling every frame independently introduces unnecessary computational overhead. We therefore employ an attention pooling module that aggregates every four consecutive frames into one temporal latent, followed by a linear projection to the ViT hidden dimension, yielding $X' \in \mathbb{R}^{(1+\frac{T}{4}) \times \frac{H}{16} \times \frac{W}{16} \times D}$.
Direct diffusion modeling in the high-dimensional DINO feature space is challenging, as informative semantic and geometric structures are often concentrated in a lower-dimensional subspace. Corrupting and denoising the full feature space requires the generative model to account for numerous directions that contribute little to the underlying representation structure, thereby increasing optimization difficulty. 
We therefore feed $X'$ into a causal ViT encoder to model channel-wise interactions and causal spatiotemporal dependencies, followed by a linear projection to obtain compact DINO-VAE latents $Z \in \mathbb{R}^{(1+\frac{T}{4}) \times \frac{H}{16} \times \frac{W}{16} \times d}$, where $d=128$. We adopt a lightweight ViT encoder to preserve the semantic and geometric structure of the original DINO features, while employing a ViT decoder to enhance reconstruction fidelity. Finally, temporal unpatchification restores the decoded representations to the original DINO feature space, yielding reconstructed features $\hat{X} \in \mathbb{R}^{(1+T) \times \frac{H}{16} \times \frac{W}{16} \times 1024}$. 

\noindent\textbf{Training Objective.} 
The raw DINO feature distribution is not explicitly regularized for diffusion-based generation. We therefore adopt a $\beta$-VAE formulation to encourage a smoother and more continuous latent distribution~\citep{higgins2017betavae}. The DINO-VAE objective is
\begin{equation}
\mathcal{L}_{\mathrm{DINO\text{-}VAE}}
=
\mathcal{L}_{\mathrm{rec}}
+
\lambda_{\mathrm{cos}}\mathcal{L}_{\mathrm{cos}}
+
\beta\mathcal{L}_{\mathrm{KL}}.
\end{equation}
Here,
$\mathcal{L}_{\mathrm{rec}} = \|\hat{X}-X\|_2^2$
is the DINO feature reconstruction loss;
$\mathcal{L}_{\mathrm{cos}} = 1 - \operatorname{Cos}(\hat{X}, X)$
is a token-wise cosine consistency loss that preserves the directional structure of DINO features; and
$\mathcal{L}_{\mathrm{KL}}
=
D_{\mathrm{KL}}\!\left(
q_{\phi}(Z \mid X)
\,\|\, 
\mathcal{N}(0,I)
\right)$
regularizes the latent distribution.
Overall, DINO-VAE compresses the high-dimensional and temporally redundant DINO features into a compact visual latent space, providing a generative space for future representation prediction and subsequent visual-flow joint modeling.

\subsection{Unified Representation and Flow Modeling} 
\label{sec:joint_modeling}

Building on the compact visual latents produced by DINO-VAE, MotionRep jointly models future visual representations and object-centric dynamics in a unified generative framework, as shown in Fig.~\ref{fig:architecture}. MotionRep employs DINO features to represent the evolution of future states, while explicitly modeling the 3D motion and geometric changes of target objects through 3D object flow.

\noindent\textbf{3D Object Flow Encoding.} 
We use the definition of 3D object flow from Sec.~\ref{sec:Overview}. Since flow is represented directly in metric 3D space, its numerical scale can differ substantially from the Gaussian noise used in diffusion-based generative modeling. We therefore normalize flow with a global scale factor $s_{\mathrm{flow}}$ as $\tilde{F}_{t,p} = \frac{F_{t,p}}{s_{\mathrm{flow}}}$, yielding a representation whose scale is better suited for diffusion-based generation. 
To associate each flow token with its corresponding geometric location, we further use the initial 3D coordinate $P_{0,p}$ as a geometric anchor. Specifically, we separately encode the global 3D position of each point and its local geometry relative to the object centroid, and fuse the two as the 3D geometric positional encoding for each flow token.

\noindent\textbf{Cross-Modal Dynamics Modeling.}
We then concatenate the representation latent sequence produced by DINO-VAE with the flow token sequence, and feed them into a unified diffusion transformer together with the clean initial object points $P_0$ for joint modeling.
Joint attention allows visual and flow tokens to interact bidirectionally: predicted future visual states can guide flow generation, while flow provides explicit object-motion information for predicting future scene states.
To further capture cross-modal spatiotemporal correspondences, we introduce spatiotemporal RoPE. We assign spatial coordinates to each flow token based on the image-plane location of its initial 3D point $P_{0,p}$ in the DINO feature map, while aligning its temporal coordinate with the corresponding DINO latent timestep, thereby establishing explicit cross-modal spatiotemporal alignment within joint attention.
Finally, separate output heads project the visual and flow representations back into their corresponding spaces, yielding the predicted future visual latents and 3D object flow.
Furthermore, the generated DINO representations can support downstream tasks such as RGB reconstruction, depth estimation, and semantic segmentation through task-specific decoders. Implementation details are provided in the Appendix.

\noindent\textbf{Modality-Specific Noise Scheduling.}
To strengthen conditional modeling between visual representations and 3D object flow, we decouple diffusion timestep sampling and noise scheduling across the two modalities. During training, we independently sample diffusion timesteps \(\tau_R\) and \(\tau_F\) for the visual representation and flow streams, respectively. The sampled timesteps are subsequently mapped to modality-specific noise levels using the shifted schedule
$S_s(\tau)=\frac{s\tau}{1+(s-1)\tau}$,
yielding $\sigma_R=S_{s_R}(\tau_R)$ and $\sigma_F=S_{s_F}(\tau_F)$. We use separate noise schedules for the two modalities to account for their different generation difficulties. 
In particular, the flow schedule is biased toward higher noise levels, allowing visual representations to form stable structures earlier and continuously guide 3D object flow generation during inference.
The forward corruption process is defined as
\begin{equation}
x_{\sigma_R}^{R}
=
(1-\sigma_R)x_{0}^{R}
+
\sigma_R\epsilon_R,
\qquad
\epsilon_R\sim\mathcal{N}(0,I).
\end{equation}
\begin{equation}
x_{\sigma_F}^{F}
=
(1-\sigma_F)x_{0}^{F}
+
\sigma_F\epsilon_F,
\qquad
\epsilon_F\sim\mathcal{N}(0,I).
\end{equation} 
Here, $x_0^R = Z_{1:K}$ and $x_0^F = \tilde{F}_{1:T}$ denote the clean future DINO-VAE latents and normalized 3D object flow, with $K=T/4$.
Independent timestep sampling exposes the model to varied visual–flow noise combinations, supporting both visual-conditioned flow generation and flow-conditioned visual prediction.

\subsection{Training and Inference}
\label{sec:training}

\noindent\textbf{Joint Training Objective.}
Both the DINO visual representation latent and 3D object flow adopt the x-prediction parameterization. For high-dimensional representation latents, meaningful structure is often concentrated on a lower-dimensional data manifold; directly predicting the clean representation therefore avoids explicitly regressing all noise directions and reduces the optimization burden. The model jointly predicts the clean states as
\begin{equation}
(\hat x_0^R,\hat x_0^F)
=
f_\theta
\left(
x_{\sigma_R}^R,x_{\sigma_F}^F,
\sigma_R,\sigma_F;\,Z_0,P_0,l
\right).
\end{equation}
The corresponding velocity is computed as
$\hat{v}
=
\frac{x_{\sigma}-\hat{x}_{0}}
{\max(\sigma,\delta)}$, where $\delta$ is a small constant for numerical stability, and the target velocity is $v=\epsilon-x_0$.

The training objective combines the velocity-matching losses of the two modalities:
\begin{equation}
\mathcal{L}
=
\lambda_R\mathcal{L}_{R}
+
\lambda_F\mathcal{L}_{F},
\end{equation}
where
\begin{equation}
\mathcal{L}_{R}=\|\hat{v}^{R}-v^{R}\|_2^2,
\quad
\mathcal{L}_{F}=\|\hat{v}^{F}-v^{F}\|_2^2.
\end{equation}

MotionRep is trained in two stages. We first pretrain the visual representation world model to learn future visual state generation, and then introduce 3D object flow into the pretrained model and jointly train both modalities.

\noindent\textbf{Unified Multi-Task Inference.}
After training, MotionRep supports multiple inference modes by configuring the noise states and denoising trajectories of the two modalities. In joint generation, the two streams share the same solver steps, while modality-specific noise schedules keep the visual representation stream ahead of flow in denoising progress, with flow remaining at a relatively higher noise level. This creates a latent-forcing-style asymmetric denoising trajectory, in which progressively refined future visual representations continuously provide informative state constraints for flow generation.

For visual-conditioned inverse dynamics, we fix the visual noise level at $\sigma_R=0$ and generate 3D object flow from noise, inferring object motion from ground-truth future visual states. Conversely, for flow-conditioned forward dynamics, we keep the given object flow clean and generate future visual representations from noise. Together with joint generation, these modes enable bidirectional conditional generation within a unified model.

\begin{table*}[t]
\centering
\caption{\textbf{3D object flow generation on WorldFlowBench.}
We evaluate 3D point trajectories over a 49-frame horizon.
ADE and FDE measure trajectory error in meters, while PWT
reports the mean point-frame hit rate averaged over thresholds
of 1\,cm, 2\,cm, and 5\,cm.}
\label{tab:worldflowbench}
\resizebox{\textwidth}{!}{%
\begin{tabular}{llccccccccc}
\toprule
\multirow{2}{*}{\textbf{Paradigm}} &
\multirow{2}{*}{\textbf{Model}} &
\multicolumn{3}{c}{\textbf{Sim7}} &
\multicolumn{3}{c}{\textbf{Real3}} &
\multicolumn{3}{c}{\textbf{SpatialCog}} \\
\cmidrule(lr){3-5}
\cmidrule(lr){6-8}
\cmidrule(lr){9-11}
& &
\textbf{ADE}$\downarrow$ & \textbf{FDE}$\downarrow$ & \textbf{PWT}$\uparrow$
& \textbf{ADE}$\downarrow$ & \textbf{FDE}$\downarrow$ & \textbf{PWT}$\uparrow$
& \textbf{ADE}$\downarrow$ & \textbf{FDE}$\downarrow$ & \textbf{PWT}$\uparrow$ \\
\midrule

\multirow{5}{*}{\shortstack[l]{Video generation\\+ 3D tracker}}
& NovaFlow
& 0.1964 & 0.3186 & 0.2897
& 0.0959 & 0.1290 & 0.2930
& 0.1657 & 0.2951 & 0.3200 \\

& Cosmos3-Nano + Track4World
& 0.1008 & 0.2021 & 0.4510
& 0.0881 & 0.1333 & 0.3095
& 0.1409 & 0.2571 & 0.3978 \\

& Cosmos3-Nano + 4RC
& 0.0963 & 0.1958 & 0.5308
& 0.0847 & 0.1219 & \underline{0.3318}
& 0.1412 & 0.2632 & 0.4012 \\

& LingBot-Video-MoE + Track4World
& 0.1051 & 0.2072 & 0.4247
& 0.0967 & 0.1453 & 0.2670
& 0.1338 & 0.2387 & 0.3644 \\

& LingBot-Video-MoE + 4RC
& 0.1035 & 0.2072 & 0.4919
& 0.0924 & 0.1344 & 0.2951
& 0.1311 & 0.2321 & 0.3755 \\

\midrule

\multirow{3}{*}{\shortstack[l]{Direct flow\\generation}}
& MolmoMotion-AR (H1)
& 0.1796 & 0.3297 & 0.2940
& 0.1818 & 0.3308 & 0.2501
& 0.2762 & 0.5331 & 0.1903 \\

& $\mu_0$
& 0.1868 & 0.3073 & 0.0698
& 0.1164 & 0.1564 & 0.1480
& 0.2132 & 0.2705 & 0.0553 \\

& EgoWAM
& 0.1987 & 0.3365 & 0.0209
& 0.1301 & 0.1901 & 0.0562
& 0.2493 & 0.3838 & 0.0800 \\

\midrule

\multirow{2}{*}{\shortstack[l]{Joint visual and\\flow generation}}
& WorldFlow
& \underline{0.0510} & \underline{0.0940} & \underline{0.5640}
& \underline{0.0810} & \underline{0.1130} & 0.3110
& \underline{0.0600} & \underline{0.0780} & \underline{0.4490} \\

& \textbf{MotionRep (ours)}
& \textbf{0.0447} & \textbf{0.0768} & \textbf{0.5965}
& \textbf{0.0657} & \textbf{0.0892} & \textbf{0.3521}
& \textbf{0.0567} & \textbf{0.0743} & \textbf{0.4503} \\

\bottomrule
\end{tabular}%
}
\end{table*}

\section{Experiments}

\subsection{Experimental Setup}

We systematically evaluate MotionRep across multiple aspects, including 3D object flow generation, DINO-VAE representation analysis, ablation studies of key design choices, and multi-task dynamics inference. 
Our experiments span simulated and real-world manipulation scenarios, with comparisons against advanced 3D object flow world models.
Further details on datasets, benchmark protocols, baseline implementations, and evaluation metrics are provided in the Appendix.

\subsection{Evaluation of Flow Generation}

\noindent\textbf{Quantitative Analysis.}
We first evaluate MotionRep on 3D object flow generation using WorldFlowBench~\citep{lu2026worldflow}. Given the initial frame, the initial 3D point set of the target object, and a task description, the model generates the full future object flow sequence. 
Table~\ref{tab:worldflowbench} compares MotionRep with video generation pipelines followed by 3D tracking, direct flow generation methods, as well as joint visual and flow generation methods.

MotionRep achieves strong trajectory prediction performance across all three benchmarks, demonstrating its ability to accurately model future object motion. On Sim7, MotionRep achieves an ADE of 0.0447~m, an FDE of 0.0768~m, and a PWT of 0.5965. It also maintains stable and competitive performance on the real-world Real3 and SpatialCog benchmarks, indicating robustness to more complex appearance variations, interaction patterns, and motion distributions. Compared with methods that jointly generate future states and flow in pixel-reconstruction-oriented VAE latent spaces, MotionRep models future dynamics in the DINO representation space, providing richer semantic and geometric state constraints for flow generation. Overall, these results show that joint modeling in a visual representation space improves the trajectory accuracy and stability of 3D object flow generation.

\begin{table}[t]
\centering

\begin{minipage}[t]{0.5\textwidth}
    \vspace{0pt}
    \centering
    \small
    \setlength{\tabcolsep}{4.3pt}
    \caption{\textbf{OOD 3D flow generation on RoboMind.}
    We evaluate generalization to unseen scenes and manipulation tasks using ADE, FDE, and mean PWT. MotionRep consistently outperforms all competing methods across the three metrics.}
    \label{tab:robomind_ood}
    \resizebox{\linewidth}{!}{%
    \begin{tabular}{lccc}
        \toprule
        \textbf{Method}
        & \textbf{ADE}$\downarrow$
        & \textbf{FDE}$\downarrow$
        & \textbf{PWT}$\uparrow$ \\
        \midrule
        NovaFlow
        & \underline{0.1171}
        & 0.2401
        & \underline{0.4981} \\
        MolmoMotion-AR (H1)
        & 0.1986
        & 0.3054
        & 0.1335 \\
        $\mu_0$
        & 0.1639
        & \underline{0.2125}
        & 0.1184 \\
        WorldFlow
        & 0.1346
        & 0.2229
        & 0.3851 \\
        \textbf{MotionRep (ours)}
        & \textbf{0.0612}
        & \textbf{0.0980}
        & \textbf{0.5257} \\
        \bottomrule
    \end{tabular}%
    }
\end{minipage}
\hfill
\begin{minipage}[t]{0.48\textwidth}
    \vspace{0pt}
    \centering
    \small
    \setlength{\tabcolsep}{2pt}
    \caption{\textbf{Inference efficiency comparison.}
    We compare inference steps and latency across 3D object flow generation methods.}
    \label{tab:inference_efficiency_main}
    \resizebox{\linewidth}{!}{%
    \begin{tabular}{@{}lcc@{}}
        \toprule
        \textbf{Inference Method}
        & \textbf{Steps}
        & \textbf{Latency (s)} \\
        \midrule
        NovaFlow & 44 & 78.44 \\
        Cosmos3-Nano + Track4World & 41 & 8.19 \\
        Cosmos3-Nano + 4RC & 41 & 8.15 \\
        LingBot-Video-MoE + Track4World & 41 & 40.47 \\
        LingBot-Video-MoE + 4RC & 41 & 40.42 \\
        MolmoMotion-AR (H1) & 40 & 183.91 \\
        WorldFlow & 40 & 13.90 \\
        \textbf{MotionRep (ours)} & \textbf{3} & \textbf{2.21} \\
        \bottomrule
    \end{tabular}%
    }
\end{minipage}

\end{table}

\noindent\textbf{Out-of-Distribution Generalization.}
We further conduct out-of-distribution (OOD) evaluation on the RoboMind dataset~\citep{hou2025robomind}, which is not used during training, with results reported in Table~\ref{tab:robomind_ood}. 
MotionRep maintains robust 3D object flow generation under substantial shifts in visual appearance, scene distributions, and language instructions, demonstrating stronger generalization than pixel-VAE-based baselines.
Notably, despite being trained on the same data as WorldFlow, MotionRep achieves better performance on the unseen RoboMind dataset, demonstrating stronger generalization to novel scenes and tasks.
These results further suggest that DINO representations reduce reliance on low-level appearance by capturing object semantics and spatial structure, providing a more transferable state representation for object-centric dynamics modeling across diverse scenes.

\noindent\textbf{Inference Efficiency.}
To evaluate the inference efficiency of MotionRep, we compare the inference steps and latency of different 3D object flow generation methods under identical hardware conditions. As shown in Table~\ref{tab:inference_efficiency_main}, MotionRep predicts 48 future frames of 3D object flow in only 2.21 seconds, achieving 6$\times$ and 83$\times$ speedups over WorldFlow and MolmoMotion, respectively, demonstrating efficient joint generation.
This efficiency advantage benefits from the properties of the DINO representation space. Unlike pixel-reconstruction-oriented VAE latents, DINO representations emphasize object-level semantics and spatial geometry rather than low-level textures and appearance details. Modeling future states in this representation space therefore reduces the need to explicitly model complex high-frequency texture variations, enabling MotionRep to achieve high-quality joint generation of future visual representations and 3D object flow with only 3 sampling steps.

\begin{figure*}[t]
    \centering
    \includegraphics[width=\linewidth]{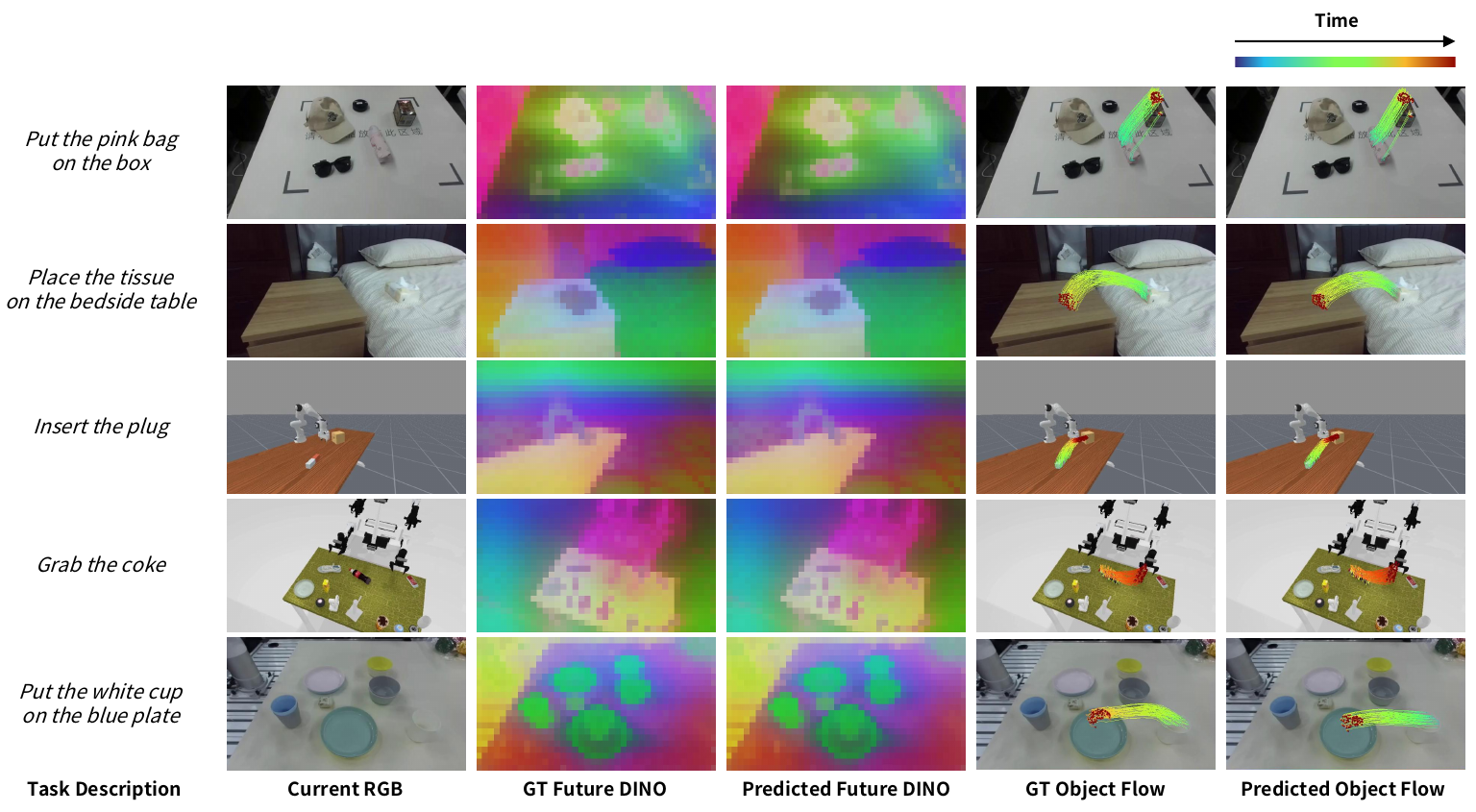}
    \caption{\textbf{Joint generation of future DINO representations and 3D object flow.}
    MotionRep jointly predicts future DINO representations and 3D object flow from the current observation and task instruction, producing future visual states and object trajectories that closely match the ground truth.}
    \label{fig:dyna_generate1}
\end{figure*}

\noindent\textbf{Qualitative Results.}
We further qualitatively evaluate the generation results of MotionRep. Fig.~\ref{fig:dyna_generate1} compares the predicted future DINO representations and 3D object flow with their ground-truth counterparts across diverse manipulation scenarios. MotionRep accurately captures object motion directions and overall trajectories, producing predictions that closely align with the ground truth in both trajectory shape and target positions across different tasks. Meanwhile, the predicted future DINO representations preserve object spatial structures and scene layouts, exhibiting strong consistency with the ground-truth representations. These results demonstrate that MotionRep effectively models future visual states and object-centric 3D motion jointly within a unified framework.

\newcommand{\metrichead}{\textbf{ADE}$\downarrow$ & \textbf{FDE}$\downarrow$ & \textbf{PWT}$\uparrow$}
\newcommand{\tbdthree}{TBD & TBD & TBD}
\newcommand{\tbdrow}[3]{& #1 & #2 & #3 & \tbdthree & \tbdthree & \tbdthree \\}

\subsection{Analysis of DINO-VAE Representations}

DINO-VAE aims to compress DINO features while preserving information essential for effective world modeling. To assess its representation quality, we conduct two complementary evaluations. First, we apply semantic probes directly to the DINO-VAE latent space to examine whether the compressed representations retain high-level semantic structure. Second, we evaluate the reconstruction quality of DINO-VAE using frozen downstream task decoders.
Additional experimental results and implementation details are provided in the Appendix.

\begin{wraptable}{r}{0.52\textwidth}
    \centering
    \small
    \setlength{\tabcolsep}{4pt}
    \renewcommand{\arraystretch}{1.0}
    \caption{\textbf{Semantic structure of the DINO-VAE latent space.}
    We evaluate semantic information preservation using top-1 classification accuracy on UCF101. Single and TTA denote single-view and test-time-augmented evaluation, respectively.}
    \label{tab:semantic_probing}

    \begin{tabular}{@{}lcc@{}}
        \toprule
        \multirow{2}{*}{\textbf{Representation}}
        & \multicolumn{2}{c}{\textbf{Top-1 Acc. (\%)}} \\
        \cmidrule(lr){2-3}
        & \textbf{Single} $\uparrow$
        & \textbf{TTA} $\uparrow$ \\
        \midrule

        DINOv3-L
        & 92.84 & 93.26 \\

        DINO-VAE reconstruction (ours)
        & 90.51 & 91.07 \\

        \midrule

        CogVideoX VAE
        & 14.51 & 14.14 \\

        Wan2.2 VAE
        & 16.94 & 16.29 \\

        AToken
        & 30.83 & 30.29 \\

        \textbf{DINO-VAE latent (ours)}
        & \textbf{79.17} & \textbf{81.79} \\

        \bottomrule
    \end{tabular}
\end{wraptable}

\noindent\textbf{Semantic Structure of the Latent Space.}
Since MotionRep performs generative modeling directly in the DINO-VAE latent space, we evaluate the semantic information retained in the latent representation. Following the DINOv3 protocol, we train a linear probe on UCF101~\citep{soomro2012ucf101} for classification. As shown in Table~\ref{tab:semantic_probing}, the original DINOv3-L features achieve a Top-1 accuracy of 92.84\%, while the compressed DINO-VAE latent achieves 79.17\%. In contrast, pixel-reconstruction-oriented video VAE latents from CogVideoX and Wan2.2 achieve only 14.51\% and 16.94\%~\citep{guo2026v}. This substantial gap indicates that DINO-VAE retains considerably richer semantic structure in its latent space. Overall, the DINO-VAE latent preserves rich semantics under strong compression, enabling effective joint modeling of visual representations and flow.

\noindent\textbf{Reconstruction Quality Evaluation.}
We evaluate the reconstruction quality of DINO-VAE using cosine similarity and MSE, and assess the preservation of geometric and semantic information in the reconstructed DINO features through depth estimation and semantic segmentation tasks.
Each decoder has fewer than 1M parameters and is trained only on original DINOv3-L features. We apply these frozen decoders to original DINO features, DINO-VAE reconstructions, and matched-budget PCA reconstructions. PCA serves as a linear compression baseline for DINO features. As shown in Table~\ref{tab:e5-decoder}, DINO-VAE reconstruction retains performance close to the original DINO features on depth estimation and semantic segmentation, while outperforming PCA reconstruction. This indicates that DINO-VAE enables high-quality reconstruction of DINO features while preserving geometric and semantic information.

\begin{table*}[t]
\centering
\small
\setlength{\tabcolsep}{5pt}
\renewcommand{\arraystretch}{1.0}
\caption{\textbf{Reconstruction quality of DINO-VAE.}
We evaluate feature similarity and the preservation of geometric and semantic information using depth and segmentation probes. DINO-VAE achieves reconstruction quality close to the original DINOv3-L features and outperforms PCA under the same compression budget.}
\label{tab:e5-decoder}

\begin{tabular*}{0.94\textwidth}{@{\extracolsep{\fill}}lccccccc@{}}
\toprule

\multirow{2}{*}{\textbf{Features}}
& \multicolumn{2}{c}{\textbf{Reconstruction}}
& \multicolumn{3}{c}{\textbf{Depth Estimation}}
& \multicolumn{2}{c}{\textbf{Segmentation}} \\

\cmidrule(lr){2-3}
\cmidrule(lr){4-6}
\cmidrule(lr){7-8}

& \textbf{Cos.} $\uparrow$
& \textbf{MSE} $\downarrow$
& \textbf{AbsRel} $\downarrow$
& \textbf{RMSE (m)} $\downarrow$
& \textbf{$\delta_1$} $\uparrow$
& \textbf{mIoU} $\uparrow$
& \textbf{IoU$_{\mathrm{Target}}$}$\uparrow$ \\

\midrule

DINOv3-L
& 1.0000
& 0.0000
& 0.0471
& 0.0987
& 0.9777
& 0.9204
& 0.6977 \\


PCA (matched budget)
& 0.8513
& 0.3973
& 0.1714
& 0.2410
& 0.6873
& 0.8753
& 0.5471 \\

\textbf{DINO-VAE (ours)}
& 0.9640
& 0.1013
& 0.0556
& 0.1086
& 0.9721
& 0.9097
& 0.6593 \\

\bottomrule
\end{tabular*}
\end{table*}

\begin{table*}[t]
    \centering
    \small
    \setlength{\tabcolsep}{8pt}
    \caption{\textbf{Ablation studies of MotionRep.}
    We evaluate the contributions of DINO-VAE, visual representation world model pretraining, future DINO generation, and modality-specific noise scheduling to 3D object flow generation. All variants exhibit degraded performance, demonstrating the effectiveness of our design choices.}
    \label{tab:ablation}
    \begin{tabular}{lccc}
        \toprule
        \textbf{Method} & \textbf{ADE}$\downarrow$ & \textbf{FDE}$\downarrow$ & \textbf{PWT}$\uparrow$ \\
        \midrule
        MotionRep w/o DINO-VAE        & 0.0727 & 0.1937 & 0.5490 \\
        MotionRep w/o pretrain        & 0.0597 & 0.1496 & 0.5500 \\
        MotionRep w/o DINO generation & 0.0517 & 0.1147 & 0.5483 \\
        MotionRep w/ shared schedule  & 0.0477 & 0.1072 & 0.5440 \\
        \textbf{MotionRep (ours)}      & \textbf{0.0315} & \textbf{0.0738} & \textbf{0.6570} \\
        \bottomrule
    \end{tabular}%
\end{table*}

\subsection{Ablation Studies}

We further conduct ablation studies on RoboTwin to assess the contribution of key components to 3D object flow generation~\citep{chen2025robotwin}, including DINO-VAE, future DINO generation, modality-specific noise scheduling, and visual representation world model pretraining, as summarized in Table~\ref{tab:ablation}. Removing \textbf{future DINO generation} and conditioning flow only on first-frame DINO features increases ADE/FDE from 0.0315$/$0.0738 to 0.0517$/$0.1147, while PWT drops from 0.6570 to 0.5483. This shows that explicitly predicting DINO representations provides temporally evolving visual states that more effectively constrain flow generation. 
Removing \textbf{DINO-VAE} and directly modeling the original DINO features also leads to performance degradation, demonstrating that the compact DINO-VAE latent space is better suited for generative modeling than high-dimensional DINO features.

Replacing \textbf{modality-specific noise scheduling} with a shared noise schedule degrades performance, indicating that asymmetric denoising enables visual representations to provide more effective guidance for flow generation.
In addition, joint training from random initialization underperforms pretrained initialization, suggesting that \textbf{pretraining future DINO generation} provides a useful dynamics prior for flow modeling. Overall, all ablations lead to performance degradation, demonstrating the complementary contributions of these components in MotionRep.

\subsection{Bidirectional Dynamics Inference}

We evaluate the bidirectional conditional inference capability of MotionRep from two complementary directions: visual-conditioned inverse dynamics and flow-conditioned forward dynamics.

\noindent\textbf{Flow-Conditioned Forward Dynamics.}
We further evaluate MotionRep's ability to generate DINO representations under flow conditioning. Given an initial visual state and the future 3D object flow, MotionRep treats the provided flow as a clean motion condition and generates the corresponding future DINO representations, thereby modeling scene evolution under the prescribed object motion.
As shown in the supplementary material, qualitative comparisons with ground-truth future DINO features show that the generated representations evolve consistently with the conditioned flow and the ground-truth scene evolution.

\begin{wraptable}{r}{0.54\textwidth}
\centering
\small
\setlength{\tabcolsep}{6pt}
\caption{\textbf{Inverse dynamics on Sim7.}
We evaluate object flow estimation from future observations using ADE, FDE, and PWT, comparing MotionRep-IDM with Track4World.}
\label{tab:inverse-dynamics}
\begin{tabular}{@{}lccc@{}}
\toprule
\textbf{Method} & \textbf{ADE} $\downarrow$ & \textbf{FDE} $\downarrow$ & \textbf{PWT} $\uparrow$ \\
\midrule
Track4World & 0.1079 & 0.1726 & 0.4636 \\
\textbf{MotionRep-IDM (ours)} & \textbf{0.0370} & \textbf{0.0849} & \textbf{0.6220} \\
\bottomrule
\end{tabular}
\end{wraptable}

\noindent\textbf{Visual-Conditioned Inverse Dynamics.}
Under the inverse dynamics setting, we condition the model on future DINO representations and generate the corresponding 3D object flow from noise to evaluate its ability to estimate 3D object motion from future visual states. 
We evaluate on Sim7 using ADE, FDE, and PWT, since it provides ground-truth 3D trajectories.
Results in Table~\ref{tab:inverse-dynamics} show that MotionRep effectively leverages future visual states to generate 3D object flow consistent with scene evolution, outperforming the 3D tracking method Track4World. We further provide qualitative results on real-world robotic manipulation data in the appendix.

\section{Conclusion}
We presented MotionRep, a visual representation world model that jointly generates future DINO representations and 3D object flow for object-centric dynamics modeling.
DINO-VAE compresses high-dimensional DINO features into compact generative latents while preserving semantic and geometric information.
Asymmetric denoising enables progressively refined future visual representations to guide 3D object flow generation.
Experiments on simulated and real-world benchmarks demonstrate that MotionRep improves 3D object flow accuracy over the baselines, generalizes to unseen scenes, and enables efficient generation with only three denoising steps.
MotionRep further supports visual-conditioned inverse dynamics and flow-conditioned forward dynamics within the same model.
Together, these results demonstrate the value of explicitly predicting future visual representations for 3D object flow generation.

\noindent\textbf{Limitations and Future Work.}
A key limitation of MotionRep is the difficulty of obtaining accurate 3D object flow annotations in real-world scenes, where supervision often relies on existing 3D trackers and may contain tracking noise. Future work will explore more reliable flow acquisition and extend MotionRep to longer-horizon tasks with more complex temporal dynamics.

\bibliography{references}
\bibliographystyle{rbs}

\clearpage
\appendix
\setcounter{figure}{0}
\setcounter{table}{0}
\renewcommand{\thefigure}{\Roman{figure}}
\renewcommand{\thetable}{\Roman{table}}
\renewcommand{\theHfigure}{appendix.\Roman{figure}}
\renewcommand{\theHtable}{appendix.\Roman{table}}

\begin{center}
    {\Large\bfseries
    Motion Beyond Pixels: Learning Future Visual Representations for Object-Centric Dynamics\par}
    \vspace{0.75em}
    {\large Supplementary Material\par}
\end{center}
\vspace{1em}

\section{Additional Method Details}

\subsection{3D Object Flow Representation}

We represent future 3D object motion using 3D object flow. Given the initial 3D point set $P_0=\{P_{0,p}\}_{p=1}^{P}$ of the target object and the corresponding point positions $P_{t,p}$ at future timestep $t$, we define flow as the cumulative displacement relative to the initial positions:
\begin{equation}
    F_{t,p} = P_{t,p} - P_{0,p},
\end{equation}
where $t=1,\ldots,T$ and $T$ denotes the prediction horizon. The resulting flow sequence is represented as $F_{1:T} \in \mathbb{R}^{T \times P \times 3}$. To normalize motion magnitudes across different scenes, we apply a global scaling factor $s_{\mathrm{flow}}$:
\begin{equation}
    \widetilde{F}_{1:T} = \frac{F_{1:T}}{s_{\mathrm{flow}}}.
\end{equation}
Unlike visual representations, the normalized flow requires no additional compression encoder and is directly perturbed with noise for generative modeling in the 3D displacement space. We then flatten the normalized flow along the temporal and point dimensions and linearly project it into the transformer hidden space.
To anchor each trajectory to its initial 3D geometry, we introduce a geometry embedding derived from the initial point positions.
For each point $p$, we encode both its absolute 3D position $P_{0,p}$ and its relative position $P_{0,p}-\bar{P}_0$, where $\bar{P}_0=\frac{1}{P}\sum_{p=1}^{P}P_{0,p}$ denotes the object centroid. These coordinates are separately processed through Fourier positional encoding and MLPs, and subsequently fused into a geometry embedding. 

Furthermore, to establish spatiotemporal alignment between flow tokens and DINO representations within the joint transformer, we employ spatiotemporal RoPE based on the source pixel locations in the initial frame. Specifically, the initial pixel coordinates corresponding to each 3D point $P_{0,p}$ are mapped onto the DINO feature grid, yielding spatial coordinates $(h_p^0,w_p^0)$. Each flow token at timestep $t$ uses temporal coordinates aligned with the temporally compressed DINO representations, together with its fixed initial spatial coordinates $(h_p^0,w_p^0)$, for RoPE encoding.

\subsection{Modality-Specific Noise Scheduling}

\begin{figure}[h]
    \centering
    \includegraphics[width=0.6\linewidth]{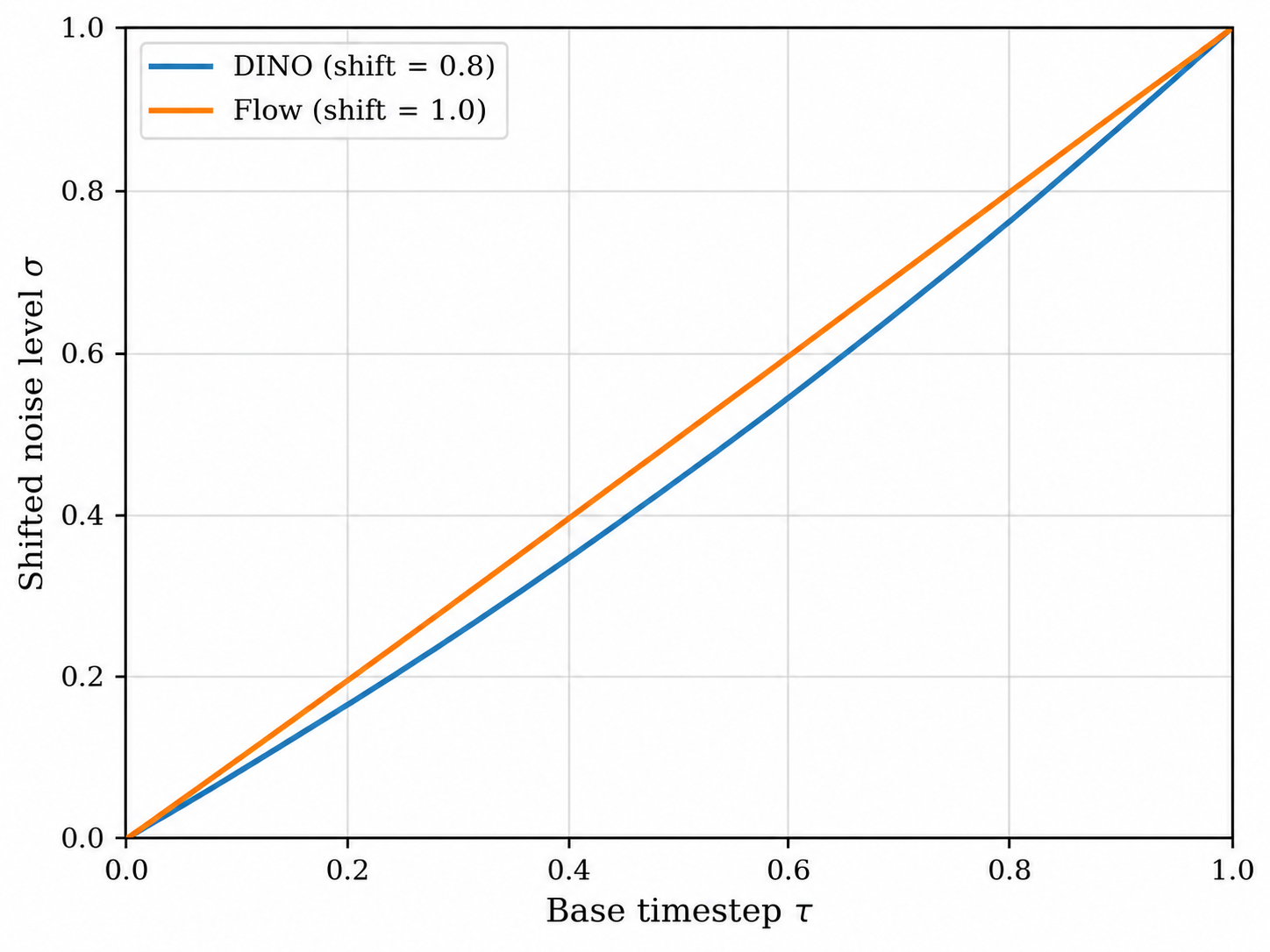}
    \caption{\textbf{Modality-specific shifted noise schedules.} During inference at 256p resolution, the DINO-VAE latents and 3D object flow use shifts of $s_R=0.8$ and $s_F=1.0$, respectively. A larger shift corresponds to a higher noise level at the same base timestep, enabling asymmetric denoising in which progressively refined visual representations guide 3D object flow generation.}
    \label{fig:shift}
\end{figure}

We provide further details on the noise scheduling strategy for visual representations and 3D object flow, as well as the asymmetric denoising process during inference. During training, we adopt Waver timestep sampling for both modalities, independently sampling timesteps $\tau_R$ and $\tau_F$. This exposes the model to diverse cross-modal noise configurations, enabling it to learn conditional dependencies across different noise levels. To account for their distinct generative characteristics, we apply modality-specific timestep shifts using
\begin{equation}
    S_s(\tau) = \frac{s\tau}{1+(s-1)\tau},
\end{equation}
yielding noise levels $\sigma_R=S_{s_R}(\tau_R)$ and $\sigma_F=S_{s_F}(\tau_F)$. During inference, we assign a larger shift to flow, allowing visual representations to maintain lower noise levels and provide more reliable structural guidance. As shown in Fig.~\ref{fig:shift}, at 256p resolution,  we set $s_R=0.8$ and $s_F=1.0$, enabling progressively refined future visual states to guide flow generation.

During training, since the number and spatial distribution of visual tokens vary with resolution, we follow the resolution-aware scheduling strategy of Cosmos3 and adjust $s_R$ accordingly to accommodate different noise characteristics. 
In contrast, we use a fixed timestep shift for 3D object flow during training, as the number of sampled points remains constant.
During inference, the visual representation and flow streams share the same solver progress but follow distinct denoising trajectories determined by their respective modality-specific schedules. This asymmetric denoising process allows progressively refined future visual states to provide continuous structural constraints for 3D object flow generation within the unified joint model.

\subsection{Unified Inference Modes}

Benefiting from independent timestep sampling across modalities during training, MotionRep supports three inference modes within a unified generative framework: joint generation, visual-conditioned inverse dynamics, and flow-conditioned forward dynamics. As summarized in Table~\ref{tab:inference_modes}, all three modes share the same model parameters and differ only in the conditioning and denoising configurations of future visual representations and 3D object flow.

\begin{table}[h]
\centering
\small
\setlength{\tabcolsep}{5pt}
\caption{\textbf{Unified inference modes of MotionRep.}
Each mode specifies whether future DINO representations and 3D object flow are generated from noise or provided as clean conditions.}
\label{tab:inference_modes}
\begin{tabular}{lcc}
\toprule
\textbf{Inference Mode} &
\textbf{Future DINO} &
\textbf{3D Object Flow} \\
\midrule
Joint Generation & Generated & Generated \\
Inverse Dynamics & Given (Clean) & Generated \\
Forward Dynamics & Generated & Given (Clean) \\
\bottomrule
\end{tabular}
\end{table}

In \textbf{joint generation}, both future visual representations and 3D object flow are initialized with Gaussian noise and jointly denoised following their respective modality-specific schedules. In \textbf{inverse dynamics}, ground-truth future DINO representations are provided as clean conditions, while only the flow stream is iteratively denoised to generate the corresponding 3D object motion. Conversely, \textbf{forward dynamics} conditions on the given clean 3D object flow and generates future visual representations consistent with the specified motion.

For conditional inference, the conditioning modality remains fixed throughout denoising, with its noise level set to $\sigma=0$. It is still assigned the corresponding timestep embedding at $\sigma=0$, preserving the timestep-conditioning mechanism used during training. Both modalities participate in joint attention, while iterative denoising is applied only to the generated modality.

\section{Implementation Details}

We provide further implementation details of MotionRep, including the architectural configurations and parameter counts of both DINO-VAE and the joint DINO-Flow DiT, as well as detailed training and inference settings.

\subsection{DINO-VAE Implementation}

The frozen DINOv3-L encoder is excluded from parameter counting. DINO-VAE comprises six trainable modules with a total of 117.08M parameters (Table~\ref{tab:dvae-params}). Temporal attention pooling compresses the $1+T$ input frames into $1+T/4$ latent frames, followed by a three-layer block-causal ViT encoder operating at hidden dimension $D=1024$. A latent head then predicts the mean $\mu$ and log-variance $\log\sigma^2$ in a $d=128$-dimensional latent space. The decoder first projects the latents back to dimension $D$, processes them through a three-layer ViT decoder, and restores the original temporal resolution through temporal unpatchification to reconstruct the input DINO features.

DINO-VAE achieves approximately $32\times$ compression through $4\times$ temporal and $8\times$ channel compression, without spatial downsampling. Attention is causal across latent frames while remaining bidirectional within each frame, ensuring that the first latent depends exclusively on the initial observation and provides a leakage-free conditioning signal. We train DINO-VAE for five epochs using AdamW with a warmup-stable-decay learning rate schedule, a peak learning rate of $10^{-4}$.

\begin{table*}[t]
\centering
\small
\setlength{\tabcolsep}{6pt}
\caption{\textbf{Trainable parameters of DINO-VAE.}
The frozen DINOv3-L encoder is excluded. The two ViT towers dominate the count and are symmetric, while the two modules that change width account for 0.4M parameters in total.}
\label{tab:dvae-params}
\begin{tabular}{@{}llr@{}}
\toprule
\textbf{Module} & \textbf{Function} & \textbf{Params} \\
\midrule
Temporal attention pooling & $(1+T) \to (1+T/4)$ along time & 3.15M \\
ViT encoder (3 layers)     & Context at width $D$ & 37.79M \\
Latent head                & $D \to d$, emits $\mu$ and $\log\sigma^2$ & 0.26M \\
Latent entry projection    & $d \to D$ & 0.13M \\
ViT decoder (3 layers)     & Context at width $D$ & 37.79M \\
Temporal unpatchify        & $(1+T/4) \to (1+T)$ along time & 37.95M \\
\midrule
\textbf{Total}             & & \textbf{117.08M} \\
\bottomrule
\end{tabular}
\end{table*}

\subsection{Joint DINO-Flow DiT Implementation}

We adapt the Cosmos3-Nano architecture, a mixture-of-transformers in which an understanding expert and a generation expert share one attention operation over one token sequence, and initialize
from its released weights. Cosmos3-Nano contains 15.18B parameters, of which 6.97B are trainable in the generation expert, while the remaining 8.21B parameters of the understanding expert are frozen.
The additional modules are lightweight, comprising a timestep embedding module, an object-flow branch, and two linear projection layers for visual representation input and output. These projections map between the 128-dimensional DINO-VAE latent space and the 4096-dimensional transformer hidden space. Both the DINOv3-L encoder and DINO-VAE remain frozen during training and are excluded from the reported parameter counts. The model configuration and key hyperparameters are summarized in Table~\ref{tab:model_config}.

\begin{table*}[t]
\centering
\small
\setlength{\tabcolsep}{6pt}
\caption{\textbf{Model configuration and hyperparameters of MotionRep.}}
\label{tab:model_config}
\begin{tabular}{@{}lll@{}}
\toprule
\textbf{Category} & \textbf{Hyperparameter} & \textbf{Value} \\
\midrule

\multirow[c]{4}{*}{Backbone}
& Backbone & Cosmos3-Nano \\
& Hidden dimension & 4096 \\
& Total parameters & 15.18B \\
& Trainable parameters & 6.97B \\

\midrule
\multirow[c]{6}{*}{Visual representation}
& Encoder & DINOv3-L \\
& DINO feature dimension & 1024 \\
& DINO-VAE ViT hidden dimension $D$ & 1024 \\
& DINO-VAE latent dimension $d$ & 128 \\
& Temporal compression & $4\times$ \\
& Channel compression & $8\times$ \\

\midrule
\multirow[c]{4}{*}{3D object flow}
& Number of points $P$ & 64 \\
& Prediction horizon $T$ & 48 \\
& Flow dimension & 3 \\
& Normalization scale $s_{\mathrm{flow}}$ & 0.1131 \\

\midrule
\multirow[c]{4}{*}{Training noise schedule}
& Timestep sampling & Waver \\
& Cross-modal timestep sampling & Independent \\
& Visual representation shift $s_R$ & 256p: 3.0; 480p: 5.0; 720p: 10.0 \\
& Flow shift $s_F$ & 5.0 \\

\midrule
\multirow[c]{3}{*}{Inference noise schedule}
& Denoising steps & 3 \\
& Visual representation shift $s_R$ & 256p: 0.8; 480p: 0.9; 720p: 0.9 \\
& Flow shift $s_F$ & 1.0 \\

\midrule
\multirow[c]{3}{*}{Objective}
& Prediction target & $x$-prediction \\
& Visual representation loss weight $\lambda_R$ & 1.0 \\
& Flow loss weight $\lambda_F$ & 1.0 \\

\bottomrule
\end{tabular}
\end{table*}

\subsection{Training Setup}

We adopt a two-stage training strategy consisting of DINO-only pretraining and joint DINO-flow training. In the first stage, we train the DINO DiT to model future visual representations without the object-flow branch. In the second stage, we initialize from the pretrained DINO DiT and introduce the object-flow branch to jointly model future visual representations and 3D object flow. We use 967,377 video clips collected from 11 data sources, grouped into three categories: Sim7, Real3, and SpatialCog, as detailed in Sec.~\ref{sec:dataset}.

To effectively leverage large-scale internet-sourced training data of varying origins and quality, we adopt a four-stage curriculum during joint DINO-flow training, progressively adjusting the sampling ratios across the three data categories. As summarized in Table~\ref{tab:training_curriculum}, Stage A primarily uses simulated data to establish basic motion modeling capabilities. Stage B increases the proportion of SpatialCog, a curated in-house dataset, to facilitate adaptation to real-world environments. Stage C emphasizes Real3, a large-scale internet-sourced dataset, to improve real-world motion modeling. Finally, Stage D increases the sampling ratio of SpatialCog for final refinement. The curriculum spans 36,000 training iterations, with stage-specific sampling ratios and learning rates.

\begin{table*}[t]
\centering
\small
\setlength{\tabcolsep}{8pt}
\caption{\textbf{Four-stage curriculum training schedule of MotionRep.}
The data sampling ratios and learning rates are adjusted across different stages of joint DINO-flow training.}
\label{tab:training_curriculum}
\begin{tabular}{@{}lcccccc@{}}
\toprule
\textbf{Stage} &
\textbf{Iterations} &
\textbf{Steps} &
\textbf{Sim}3 &
\textbf{Real3} &
\textbf{SpatialCog} &
\textbf{Learning Rate} \\
\midrule
A: Sim3 initialization  & 0--8,000       & 8,000  & 80\% & 10\% & 10\% & $1\times10^{-4}$ \\
B: SpatialCog emphasis      & 8,000--18,000  & 10,000  & 20\% & 10\% & 70\% & $1\times10^{-4}$ \\
C: Real3 emphasis      & 18,000--30,000 & 12,000 & 15\% & 70\% & 15\% & $8\times10^{-5}$ \\
D: SpatialCog refinement    & 30,000--36,000 & 6,000  & 10\% & 35\% & 55\% & $2\times10^{-5}$ \\
\bottomrule
\end{tabular}
\end{table*}

We train the joint model on 64 NVIDIA H100 GPUs, with approximately 40,000 iterations of DINO-only pretraining followed by 36,000 iterations of joint DINO-flow training. We use the AdamW optimizer with a global batch size of 64 packed sequences, each containing at most 55,296 tokens. An exponential moving average (EMA) of the model parameters is maintained throughout training, and the EMA weights are used for evaluation.

\subsection{Inference Setup}

During inference, we use the EMA weights of the jointly trained model to generate future visual representations and 3D object flow, conditioned on the initial observation, 3D object query points, and language instruction. We employ the UniPC solver with 3 sampling steps and a classifier-free guidance (CFG) scale of 1.2. To facilitate few-step inference, we adopt smaller noise shifts to provide more balanced coverage of the denoising trajectory. We further set the visual representation shift lower than the flow shift, enabling progressively refined future visual states to guide 3D object flow generation. Finally, the generated visual latents are decoded into DINO features using the frozen DINO-VAE decoder, while the predicted 3D object flow is rescaled and added to the initial 3D point coordinates to recover future object trajectories.

\section{Experimental Details}
\label{sec:experimental_details}

\subsection{Datasets and Benchmarks}
\label{sec:dataset}

\paragraph{Datasets.}
We collect a total of 967,377 video clips from 11 data sources, spanning simulated environments, real-world internet videos, and in-house recordings. We organize the data into three groups: \textbf{Sim7}, comprising seven simulation datasets (RoboTwin, Meta-World, ManiSkill, RoboCasa, MIKASA, RLBench, and MetaFold); \textbf{Real3}, comprising three real-world datasets (AgiBot World, Something-Something v2, and EPIC-KITCHENS); and \textbf{SpatialCog}, an in-house real-world dataset with calibrated camera parameters. The detailed data composition is summarized in Table~\ref{tab:train-mixture}. To prevent data leakage, we exclude evaluation clips from the training set.

For 3D object flow supervision, Sim7 directly uses ground-truth 3D motion provided by simulators, while 3D object trajectories in Real3 and SpatialCog are estimated from RGB videos using SAM 2 and Track4World. Specifically, we first segment the target object using SAM 2 and sample query points on its visible surface in the initial frame. We then track these points using Track4World and transform the reconstructed 3D points and their trajectories into the initial camera coordinate system using depth, camera intrinsics, and camera poses. All samples are associated with undistorted pinhole camera intrinsics. Sim7 provides ground-truth 3D geometry from simulators, SpatialCog provides calibrated camera parameters, and Real3 primarily relies on monocular estimation for 3D geometry.

Beyond 3D object flow generation, we evaluate the representation capabilities of DINO-VAE on additional datasets. For depth estimation and semantic segmentation, we use held-out robotic manipulation videos. For video representation evaluation, we use UCF101 following the evaluation protocol of DINOv3.

\paragraph{WorldFlowBench.}
We evaluate the 3D object flow generation capabilities of MotionRep on WorldFlowBench~\citep{lu2026worldflow}, which consists of 1,678 object interaction and manipulation clips from simulated and real-world environments. The benchmark is divided into three evaluation groups: Sim7 (396 clips), Real3 (882 clips), and SpatialCog (400 clips). Each evaluation sample contains an initial RGB frame at a resolution of $288\times480$, a natural-language task description, and 64 object query points specified by their pixel coordinates in the initial frame. The corresponding 3D trajectories span 49 sampled frames, including the initial frame and 48 future frames.

\begin{table}[h]
\centering
\small
\caption{\textbf{Training data composition.}
Number of video clips from each individual data source, grouped into Sim7, Real3, and SpatialCog. All clip counts are reported prior to the exclusion of evaluation samples and the subsequent train/validation split.}
\label{tab:train-mixture}
\begin{tabular}{@{}llr@{}}
\toprule
\textbf{Source} & \textbf{Group} & \textbf{Clips} \\
\midrule
AgiBot World~\citep{agibot2025world} & Real3 & 597{,}272 \\
Something-Something~v2~\citep{goyal2017something} & Real3 & 84{,}153 \\
EPIC-KITCHENS~\citep{damen2022rescaling} & Real3 & 10{,}519 \\
\midrule
RoboTwin~\citep{mu2025robotwin} & Sim7 & 115{,}775 \\
Meta-World~\citep{yu2019metaworld} & Sim7 & 52{,}477 \\
ManiSkill~\citep{tao2024maniskill3} & Sim7 & 9{,}033 \\
RoboCasa~\citep{nasiriany2024robocasa} & Sim7 & 6{,}616 \\
MIKASA~\citep{cherepanov2025mikasa} & Sim7 & 4{,}599 \\
RLBench~\citep{james2020rlbench} & Sim7 & 3{,}137 \\
MetaFold~\citep{chen2025metafold} & Sim7 & 2{,}013 \\
\midrule
Own capture        & SpatialCog  & 81{,}783 \\
\midrule
Total              &             & 967{,}377 \\
\bottomrule
\end{tabular}
\end{table}

\subsection{Evaluation Metrics}

\paragraph{3D Object Flow.}
We evaluate 3D object flow generation using Average Displacement Error (ADE), Final Displacement Error (FDE), and Percentage Within Threshold (PWT). ADE measures the mean 3D Euclidean position error across all query points and timesteps, while FDE measures the mean position error at the final timestep. PWT measures the fraction of point-frame predictions within distance thresholds of 1, 2, and 5 cm, averaged across the three thresholds. ADE and FDE are reported in meters. All metrics are computed over the prediction horizon and then averaged across clips within each evaluation group.

\paragraph{Depth and Semantic Segmentation.}
For depth estimation, we report Absolute Relative Error (AbsRel), Root Mean Squared Error (RMSE), and $\delta_1$. AbsRel measures the mean absolute relative depth error, RMSE measures the root mean squared depth error, and $\delta_1$ denotes the fraction of pixels with a predicted-to-ground-truth depth ratio within a factor of 1.25. For semantic segmentation, we report mean Intersection over Union (mIoU), averaged across five semantic classes.

\paragraph{Visual Representation Evaluation.}
Following the DINOv3 evaluation protocol, we assess the semantic structure of visual representations using top-1 classification accuracy on UCF101. We report results under two inference settings: Single and test-time augmentation (TTA). Single uses a deterministic temporal sampling strategy covering the first and last frames, with uniformly spaced intermediate frames, and a central square crop. TTA averages the predicted class probabilities across six views, constructed from three spatial crops and two temporally interleaved frame sequences.

\subsection{Baselines}

We compare MotionRep against three categories of baselines: video-generation-and-tracking pipelines that estimate 3D motion from generated videos, direct-flow methods that predict future 3D trajectories without explicit RGB video generation, and joint visual-flow generation methods that model future visual states and 3D motion together.
All methods are evaluated under the same conditioning setting, using a single initial observation, a natural-language task description, and initial 3D query points.

\paragraph{Video Generation and 3D Tracking.}

{NovaFlow}~\citep{li2026novaflow} generates future videos conditioned on the initial observation and task description, and subsequently extracts actionable 3D object flow using off-the-shelf perception modules.
{Cosmos3-Nano + Track4World} combines Cosmos3-Nano~\citep{nvidia2026cosmos3} for future RGB video generation with Track4World~\citep{lu2026track4world}, a feed-forward world-centric 3D tracker, to recover 3D point trajectories.
{Cosmos3-Nano + 4RC} replaces Track4World with 4RC~\citep{luo20264rc}, which reconstructs 3D geometry and motion from generated monocular videos.
{LingBot-Video-MoE + Track4World} uses LingBot-Video-MoE~\citep{ma2026lingbotvideo} for autoregressive video generation, followed by Track4World for 3D trajectory estimation.
{LingBot-Video-MoE + 4RC} similarly combines LingBot-Video-MoE with 4RC for 3D motion reconstruction.
These methods estimate 3D motion through an intermediate video-generation stage, making their performance dependent on both video generation and subsequent 3D tracking or reconstruction.

\paragraph{Direct Flow Generation.}
{MolmoMotion-AR (H1)}~\citep{zhang2026molmomotion} autoregressively predicts language-conditioned 3D point trajectories by representing coordinates as discrete tokens.
{$\mu_0$}~\citep{lee2026mu0} directly models future trajectories of semantic interaction points in a normalized, anchor-relative 3D trace space.
{EgoWAM}~\citep{li2026egowam} models future dynamics beyond pixels, including a 3D-flow variant that predicts camera-stabilized 3D motion.
Unlike video-generation-and-tracking pipelines, these methods directly predict geometric motion without generating RGB videos.

\paragraph{Joint Visual and Flow Generation.}
{WorldFlow}~\citep{lu2026worldflow} jointly models future RGB video and 3D object flow within a unified generative framework, serving as the most directly comparable baseline to MotionRep.

\subsection{DINO-VAE Evaluation Protocols}
\label{sec:dino_vae_details}

\paragraph{Latent Semantic Probing.}
To assess the semantic information retained in compressed DINO-VAE latents, we follow the DINOv3 evaluation protocol and perform linear probing on UCF101. Specifically, we apply spatial and temporal pooling to the video representations extracted by each model to obtain video-level feature vectors, which are then passed through LayerNorm and a single linear classifier for video classification. The classifier is trained for 20 epochs using cross-entropy loss. Under the same evaluation protocol, we compare the original DINOv3-L features, DINO-VAE latents, and video VAE latents from CogVideoX and Wan2.2 to evaluate their ability to preserve high-level semantic information.

\paragraph{Depth Decoder.}
We evaluate the quality of reconstructed DINO features and compressed DINO-VAE latents using MSE, cosine similarity, and frozen downstream decoder to assess the preservation of geometric and semantic information. The depth decoder is a lightweight convolutional network with 0.63M parameters, taking only the last-layer DINO feature as input. It constructs four feature scales from a single feature and progressively fuses them through top-down upsampling, producing a single-channel positive depth map at the input image resolution. The training objective combines a disparity-space data term, a scale-invariant logarithmic loss, a gradient loss, and an edge loss, with larger weights assigned to the latter two to improve depth boundary preservation.

\paragraph{Segmentation Decoder.}
The segmentation decoder employs a lightweight decoder with 0.76M parameters. It constructs multi-scale features from the same DINO feature through four parallel projection branches and fuses them to produce per-pixel semantic predictions. The segmentation task comprises five classes: background, table, robot, clutter, and the currently manipulated object. The training objective combines equally weighted cross-entropy and Dice losses, with inverse-frequency class weights computed from the training split to mitigate class imbalance.

\paragraph{RGB Decoder.}
We adopt a lightweight ViT RGB decoder~\citep{zheng2026rae}, initialized from a pretrained RAE pixel decoder and subsequently fine-tuned on our data, to map the last-layer DINO features back to pixel space. When reconstructing future RGB frames, we further introduce a lightweight first-frame reference module that leverages high-frequency information from the initial RGB frame to recover texture details not fully preserved in DINO features, thereby improving the visual quality of reconstructed videos.

\paragraph{PCA Visualization.}
We apply principal component analysis (PCA) to visualize the spatial structure of DINO-VAE latents. Specifically, we project the latent features onto the first three principal components and map them to the RGB channels to obtain spatial feature visualizations. Regions with similar colors correspond to more similar representations in the projected feature space, providing a qualitative view of the spatial structure preserved after temporal and channel compression.

\subsection{Out-of-Distribution Evaluation Protocol}

To further assess the out-of-distribution (OOD) generalization of MotionRep, we evaluate it on RoboMIND 2.0, a dataset entirely excluded from training. This dataset encompasses diverse manipulation scenes and tasks unseen during training, enabling us to assess the model's ability to predict 3D object motion under distribution shifts. We select 200 video clips from RoboMIND 2.0 and follow the same evaluation protocol, metrics, and settings as WorldFlowBench. The results are reported in main paper.

\subsection{Inference Efficiency Details}

To ensure a fair and reproducible evaluation of inference efficiency, we conduct all measurements on an NVIDIA H800 GPU with a batch size of 1. We select 50 samples from each of the three WorldFlowBench evaluation groups, Sim7, Real3, and SpatialCog, and use the same samples to evaluate all methods. We compare inference efficiency against several baseline methods that achieve strong performance in 3D object flow generation.

\section{Additional Experimental Results}

In this section, we provide additional experimental results of MotionRep, including DINO-VAE representation analysis, downstream evaluations of reconstructed DINO features, joint generation of future DINO representations and 3D object flow, and bidirectional dynamics inference.

\subsection{DINO-VAE Representation Analysis}

\subsubsection{PCA Visualization of Latent Representations}

\begin{figure*}[h]
    \centering
    \includegraphics[width=0.7\linewidth]{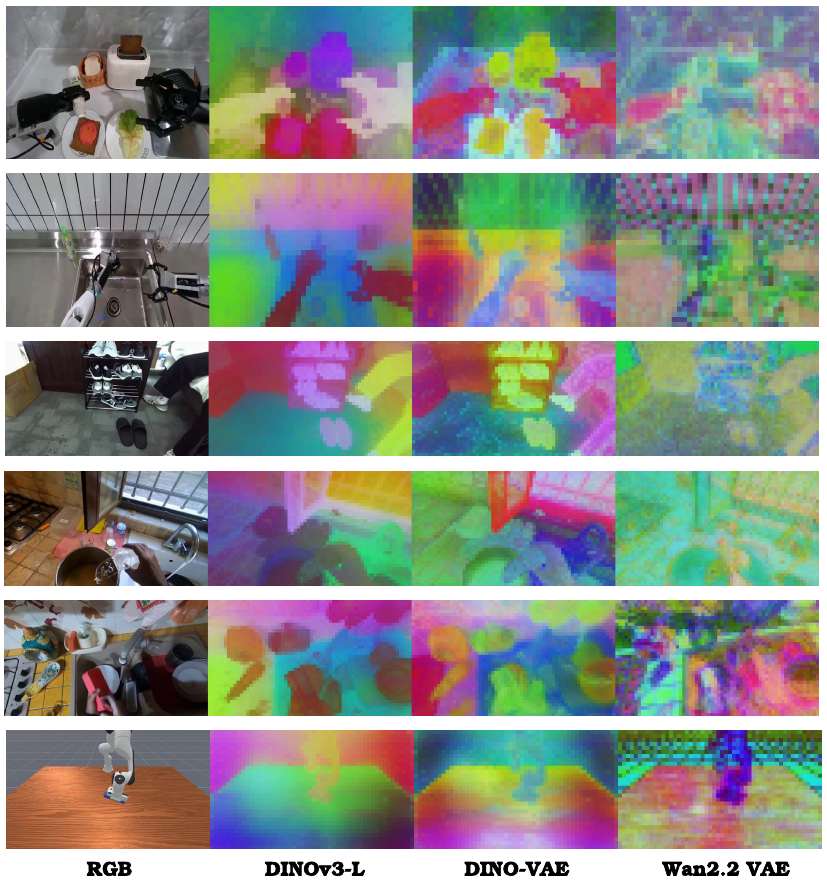}
    \caption{\textbf{PCA visualization of representation structure.}
    We map the first three principal components of DINOv3-L features, compressed DINO-VAE latents, and Wan2.2 VAE latents to RGB channels. Regions with similar colors indicate similar feature representations in the projected space. DINO-VAE largely preserves the object-level semantic and geometric organization of the original DINO features, while Wan2.2 VAE exhibits a more entangled feature structure with weaker separation between objects and background.}
    \label{fig:dinovae-latent}
\end{figure*}

We compare the spatial structures of original DINOv3-L features, compressed DINO-VAE latents, and Wan2.2 VAE latents using PCA visualization to assess their preservation of semantic and geometric information. As shown in Fig.~\ref{fig:dinovae-latent}, the original DINOv3-L features exhibit clear object-level spatial structures, with distinct feature regions corresponding to different objects and foreground-background boundaries. Despite temporal and channel compression, DINO-VAE latents largely preserve this spatial organization, exhibiting coherent feature distributions with strong object-level consistency. In contrast, Wan2.2 VAE latents exhibit more fine-grained noise-like variations and fragmented spatial patterns, with less distinct object boundaries and regional structures.

These observations suggest that DINO-VAE preserves the informative semantic and spatial structures of the original DINO features while substantially compressing their dimensionality. 
Compared with pixel-oriented VAE latents containing substantial local high-frequency variations, such coherent and structured representations allow generative models to focus on object-level semantics, spatial relationships, and dynamic changes, thereby improving inference efficiency and generalization. 
Together with the ablation results in the main paper, these findings further support the effectiveness of DINO-VAE as a latent space for jointly generating future visual representations and 3D object flow.

\subsubsection{Geometric Information in DINO-VAE Latents}

To evaluate the geometric information preserved by DINO-VAE, we independently train lightweight depth decoders on the original DINOv3-L features and compressed DINO-VAE latents, and assess their depth estimation performance using AbsRel, RMSE, and $\delta_1$. As shown in Table~\ref{tab:geometric_probing}, despite temporal and channel compression, DINO-VAE latents achieve performance close to the original DINO features. These results demonstrate that DINO-VAE effectively preserves geometric information under substantial compression, providing a compact representation space for future state generation.

\begin{table}[h]
    \centering
    \small
    \setlength{\tabcolsep}{1pt}
    \renewcommand{\arraystretch}{1.0}
    \caption{\textbf{Geometric information preservation in DINO-VAE latents.}
    Despite substantial temporal and channel compression, DINO-VAE latents preserve rich geometric information.}
    \label{tab:geometric_probing}

    \begin{tabular}{@{}lccc@{}}
        \toprule
        \multirow{1}{*}{\textbf{Representation}}
        & \textbf{AbsRel} $\downarrow$
        & \textbf{RMSE (m)} $\downarrow$
        & \textbf{$\delta_1$} $\uparrow$ \\
        \midrule

        DINOv3-L~\citep{simeoni2025dinov3}
        & 0.0423 & 0.1422 & 0.9705 \\



        \textbf{DINO-VAE latent (ours)}
        & 0.0426 & 0.1432 & 0.9668 \\

        \bottomrule
    \end{tabular}
\end{table}

\subsubsection{Reconstruction Quality and Downstream Tasks}
To further evaluate the reconstruction quality of DINO-VAE and its ability to support diverse downstream visual tasks, we apply the reconstructed DINO features to depth estimation, semantic segmentation, and RGB reconstruction. 

\begin{figure*}[h]
    \centering
    \includegraphics[width=0.7\linewidth]{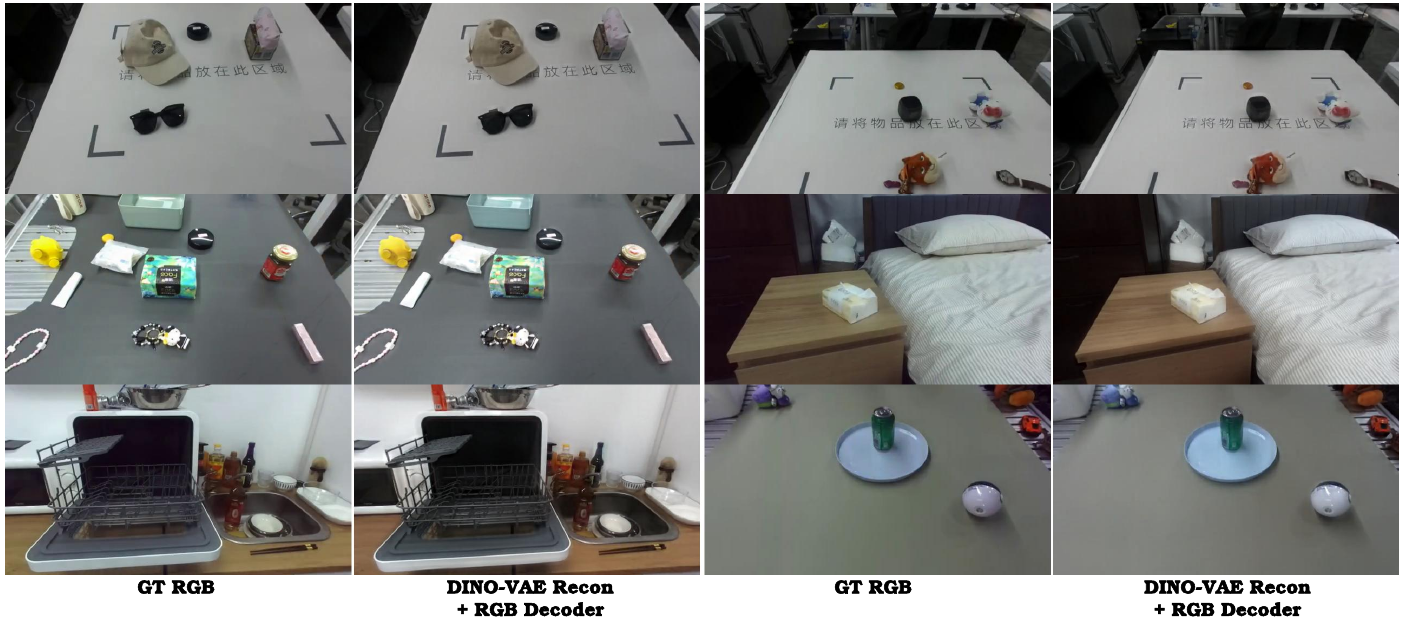}
    \caption{\textbf{RGB reconstruction from DINO-VAE reconstructed features.}
    We compare ground-truth RGB images with RGB reconstructions obtained by passing DINO-VAE reconstructed DINO features through an RGB decoder. The reconstructed images preserve the overall scene structure, object appearance, and spatial details of the original observations.}
    \label{fig:rgb-viz}
\end{figure*}

\paragraph{RGB Reconstruction.}
We decode reconstructed DINO features into RGB images and conduct both quantitative and qualitative evaluations. As shown in Table~\ref{tab:rgb_reconstruction}, the reconstructed images achieve a PSNR of 30.28 dB, an SSIM of 0.9536, and an LPIPS of 0.1375. As illustrated in Fig.~\ref{fig:rgb-viz}, the reconstructions exhibit strong consistency with ground-truth RGB frames in scene layout, object appearance, and texture details. These results demonstrate that DINO-VAE reconstructed features support high-quality RGB reconstruction.

\begin{figure*}[h]
    \centering
    \includegraphics[width=0.7\linewidth]{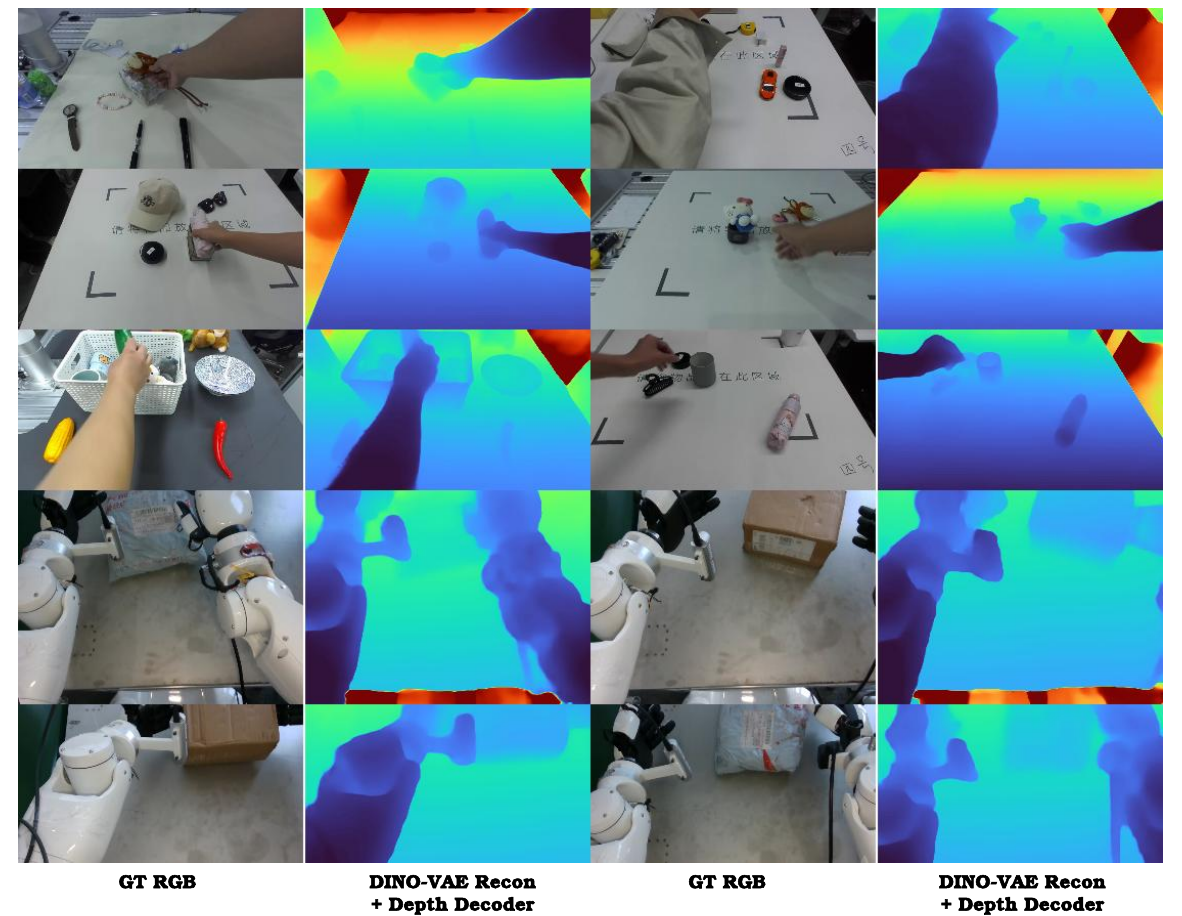}
    \caption{\textbf{Depth estimation from DINO-VAE reconstructed features.}
    The lightweight frozen depth decoder captures scene geometry and relative depth relationships between objects and their surroundings.}
    \label{fig:Dep-viz}
\end{figure*}

\paragraph{Depth Estimation.}
As shown in Fig.~\ref{fig:Dep-viz}, the predicted depth maps preserve the overall spatial structure of different manipulation scenes and distinguish relative depth relationships among robots, manipulated objects, and background regions. Across robotic manipulation and tabletop interaction scenarios, the depth predictions capture the spatial arrangement of major objects and their geometric relationships with the surrounding environment. These results provide further qualitative evidence that DINO-VAE reconstructed features retain useful geometric information for dense depth estimation.

\begin{figure*}[t]
    \centering
    \includegraphics[width=0.7\linewidth]{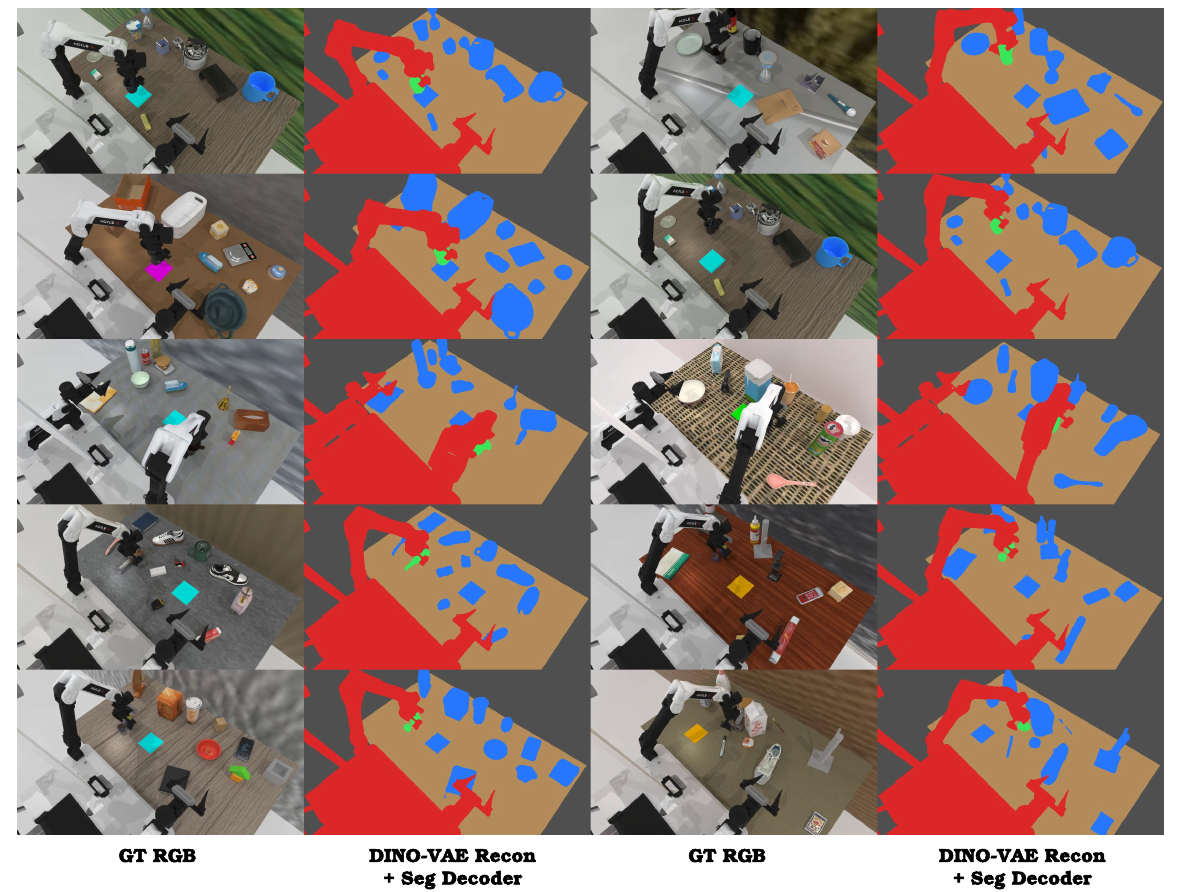}   
    \caption{\textbf{Semantic segmentation from DINO-VAE reconstructed features.}
    The lightweight frozen decoder preserves clear object-level semantic regions across diverse manipulation scenes.
    Gray, brown, red, blue, and green denote the background, table, robot, clutter, and manipulated object, respectively.}

    \label{fig:Sep-viz}
\end{figure*}

\paragraph{Semantic Segmentation.}
As shown in Fig.~\ref{fig:Sep-viz}, the predicted segmentation maps distinguish major semantic regions, including tables, robots, background, and manipulated objects, while maintaining relatively clear object-level boundaries across diverse manipulation scenarios. Together with the quantitative results in the main paper, these visualizations further demonstrate that DINO-VAE reconstructed features preserve spatial structures and object-level semantic information useful for dense semantic prediction.

\begin{table}[h]
    \centering
    \caption{\textbf{RGB reconstruction quality of DINO-VAE.}
    We evaluate RGB images decoded from DINO-VAE reconstructed features using PSNR, SSIM, and LPIPS.}
    \label{tab:rgb_reconstruction}
    \small
    \setlength{\tabcolsep}{2pt}
    \begin{tabular}{@{}lccc@{}}
        \toprule
        \textbf{Method} & \textbf{PSNR (dB)} $\uparrow$ & \textbf{SSIM} $\uparrow$ & \textbf{LPIPS} $\downarrow$ \\
        \midrule
        \textbf{DINO-VAE Reconstruction} & 30.28 & 0.9536 & 0.1375 \\
        \bottomrule
    \end{tabular}
\end{table}

\subsection{Joint Generation Results}

\begin{figure*}[h]
    \centering
    \includegraphics[width=\linewidth]{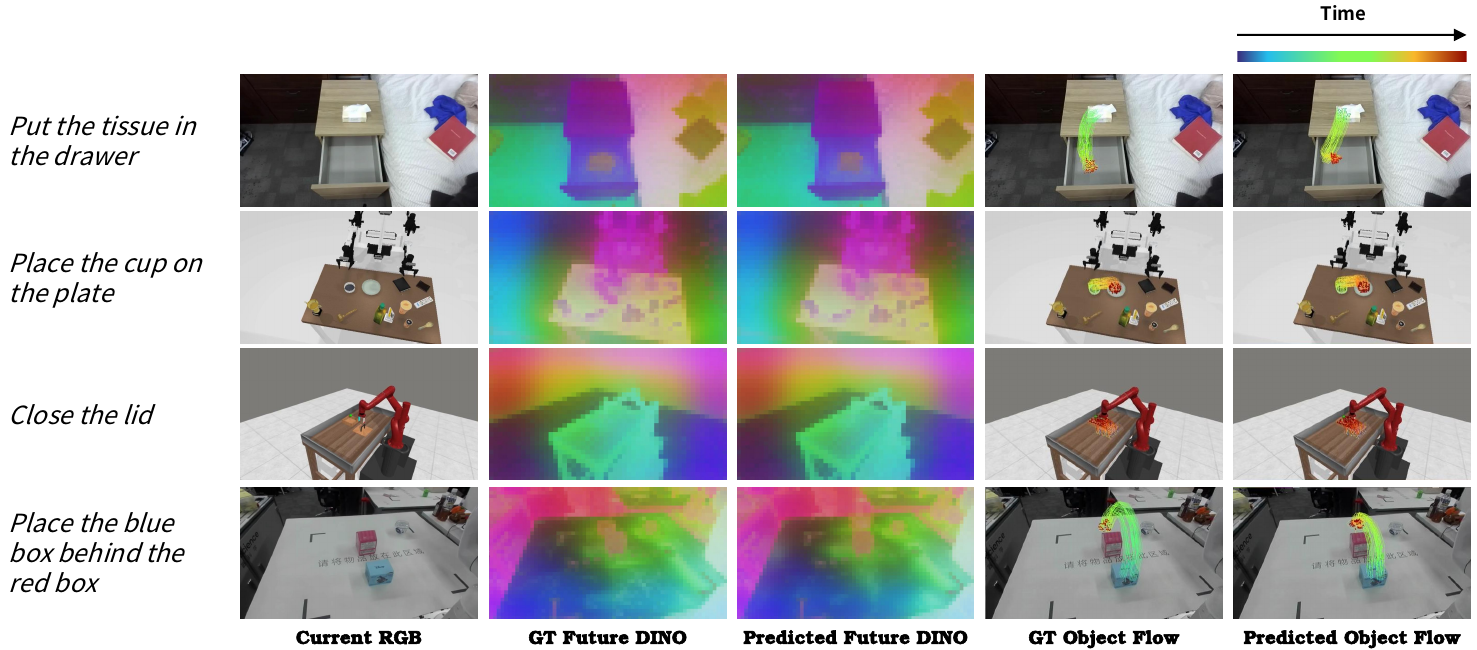}
    \caption{\textbf{Joint generation of 3D object flow and future DINO representations.}
    MotionRep jointly predicts future DINO representations and 3D object flow, producing visual states and object trajectories consistent with the ground truth.}
    \label{fig:Generated2}
\end{figure*}

We provide additional qualitative results of jointly generated future DINO representations and 3D object flow across diverse manipulation scenarios. As shown in Fig.~\ref{fig:Generated2}, 
MotionRep produces future DINO representations that closely match the ground truth in overall scene layout, object-level structures, and spatial changes across diverse manipulation tasks, including object placement, drawer interactions, and lid closing. Meanwhile, the predicted 3D object flow captures the primary motion directions and trajectory evolution of manipulated objects, with motion patterns and final positions closely aligned with the ground truth.

\subsection{Bidirectional Dynamics Inference}

Beyond joint generation, we further provide qualitative results for visual-conditioned inverse dynamics and flow-conditioned forward dynamics to examine the bidirectional conditional dependencies between future visual states and object motion.

\begin{figure*}[h]
    \centering
    \includegraphics[width=\linewidth]{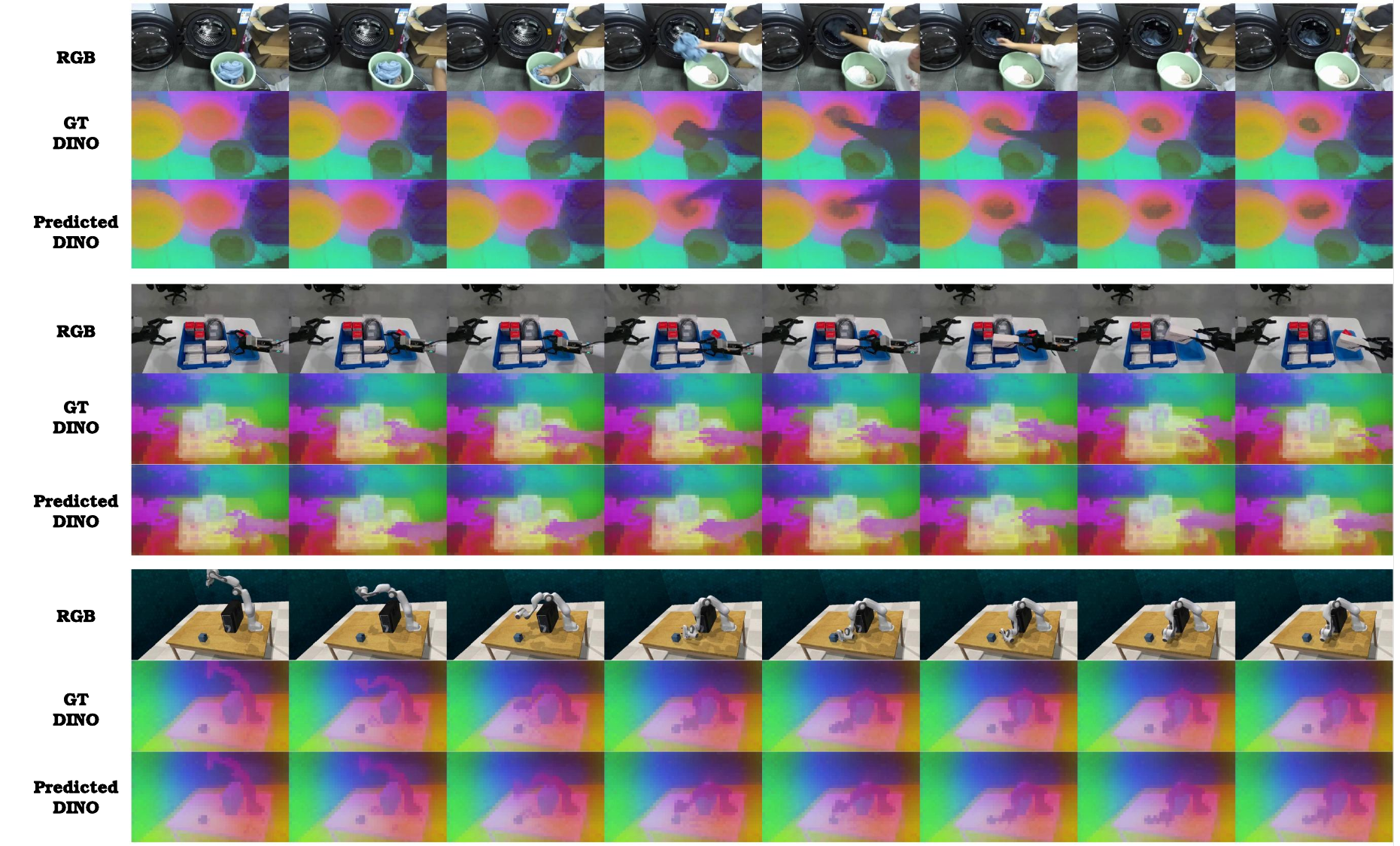}
    \caption{\textbf{Visualization of flow-conditioned forward dynamics.}
    Given ground-truth 3D object flow, MotionRep generates future DINO representations consistent with the corresponding scene evolution.}
    \label{fig:Forward-viz}
\end{figure*}

\paragraph{Flow-Conditioned Forward Dynamics.}
For forward dynamics inference, we provide ground-truth 3D object flow as a condition and generate the corresponding future DINO representations. As illustrated in Fig.~\ref{fig:Forward-viz}, the generated representations remain consistent with the ground-truth features in overall spatial layout, major object structures, and scene evolution. As the object moves, the generated DINO features capture its spatial changes in a manner consistent with the ground-truth DINO features, while preserving the relatively stable background structure.

\begin{figure*}[h]
    \centering
    \includegraphics[width=\linewidth]{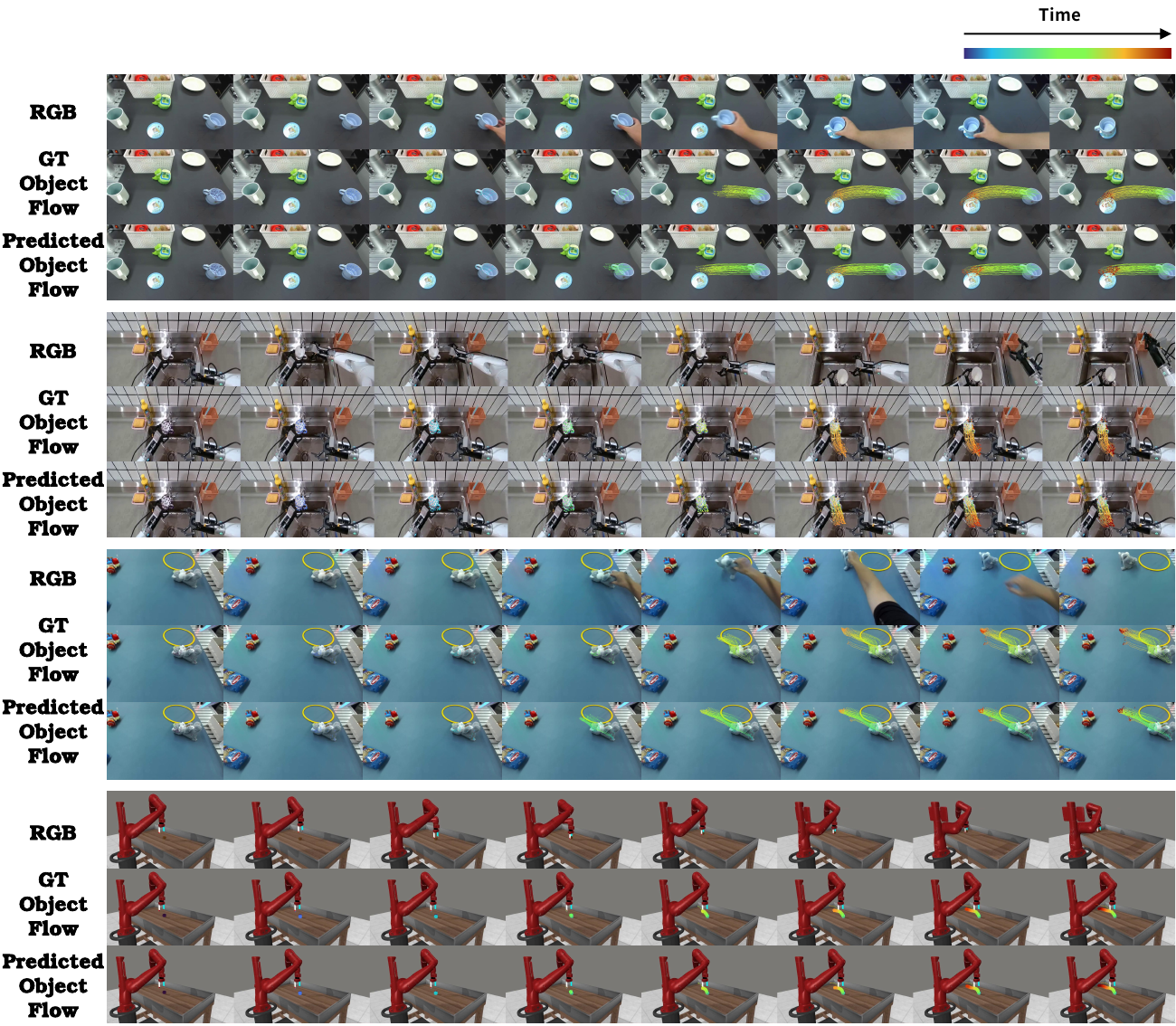}
    \caption{\textbf{Visualization of visual-conditioned inverse dynamics.}
    Given ground-truth future DINO representations, MotionRep predicts 3D object flow that closely follows the ground-truth trajectories.}
    \label{fig:Inverse-viz}
\end{figure*}

\paragraph{Visual-Conditioned Inverse Dynamics.}
For inverse dynamics inference, we provide ground-truth future DINO representations as conditions and generate the corresponding 3D object flow. As shown in Fig.~\ref{fig:Inverse-viz}, the predicted trajectories exhibit strong agreement with the ground truth in motion direction, temporal evolution, and overall displacement across real-world and simulated manipulation scenarios. In scenes involving object movement and robotic interactions, MotionRep effectively infers 3D motion from the spatial changes encoded in future visual representations. These results indicate that future DINO representations not only describe scene states but also provide informative conditions for object-centric motion inference.

\end{document}